\documentclass{article}
\usepackage{microtype}
\usepackage{graphicx}
\usepackage{subcaption}
\usepackage{booktabs} 

\usepackage{hyperref}

\usepackage[preprint]{icml}

\usepackage{amsmath}
\usepackage{amssymb}
\usepackage{mathtools}
\usepackage{amsthm}

\usepackage{algorithm}
\usepackage{algorithmic}
\usepackage{tabularx}

\usepackage[capitalize,noabbrev]{cleveref}

\theoremstyle{plain}

\theoremstyle{definition}

\theoremstyle{remark}

\icmltitlerunning{Self-Conditioned Denoising for Atomistic Representation Learning}

\begin{document}

\twocolumn[
  \icmltitle{Self-Conditioned Denoising for Atomistic Representation Learning}

\icmlsetsymbol{equal}{*}

  \begin{icmlauthorlist}
    \icmlauthor{Tynan J. Perez}{yyy}
    \icmlauthor{Luis Carretero López}{eth}
    \icmlauthor{Bowen Deng}{xxx}
    \icmlauthor{Rafael Gómez-Bombarelli}{xxx}
\end{icmlauthorlist}

  \icmlaffiliation{yyy}{Department of Chemistry, Massachusetts
Institute of Technology, 77 Massachusetts Avenue, Cambridge, 02139,
MA, USA}
  \icmlaffiliation{xxx}{Department of Materials Science and Engineering, Massachusetts
Institute of Technology, 77 Massachusetts Avenue, Cambridge, 02139,
MA, USA}
  \icmlaffiliation{eth}{ETH Zürich, Zürich, Switzerland}

  \icmlcorrespondingauthor{Rafael Gómez-Bombarelli}{rafagb@mit.edu}

\icmlkeywords{self-supervised learning, pre-training, denoising, atomistic representation learning, molecular property prediction}

  \vskip 0.3in
]

\printAffiliationsAndNotice{}

\begin{abstract}

The success of large-scale pre-training in language and vision has inspired the development of foundation models for the physical sciences; however, label-free pre-training methods for atomistic data remain under-investigated. To date, supervised pre-training on DFT force-energy labels has provided the largest benefit to downstream property prediction, outperforming self-supervised learning (SSL) methods that remain limited to ground-state geometries or single domains of atomistic data. We address these shortcomings with Self-Conditioned Denoising (SCD), a backbone-agnostic reconstruction objective in which the model's own embedding of a clean geometry, its self-embedding, conditions the denoising of a corrupted copy. SCD applies to any atomistic domain, including small molecules, proteins, periodic materials, and non-equilibrium geometries. With backbone and pre-training data matched, SCD reduces error by 20--45\% relative to standard node denoising on QM9 targets, outperforming prior denoising variants on most targets, and matches the performance of supervised force-energy pre-training on identical geometries. A small, fast GNN pre-trained by SCD is competitive with or superior to larger models pre-trained on far larger labeled or unlabeled datasets across multiple domains. We also fine-tune leading foundation potentials (MACE-OFF, Orb-v3, UMA-S) for property prediction; SCD models remain competitive with them at one to two orders of magnitude less pre-training compute. Our code is available at: \url{https://github.com/TyJPerez/SelfConditionedDenoisingAtoms}

\end{abstract}

\begin{figure}[t]
\centering
\includegraphics[width=\columnwidth]{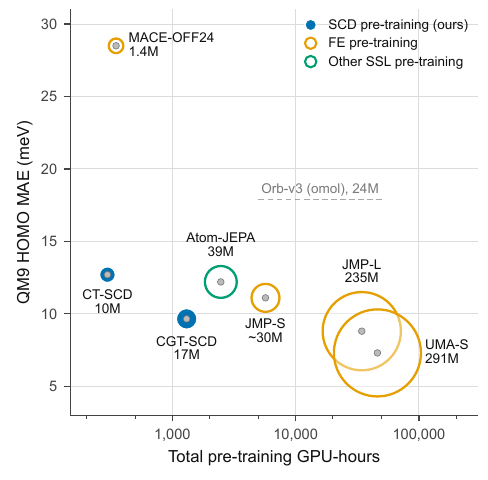}
\caption{\textbf{Pre-training compute vs.\ QM9 HOMO error.} Marker area is proportional to parameter count (the MACE-OFF24 marker is enlarged for legibility). Filled markers: our SCD pre-trained models (CT and CGT); orange rings: force-energy pre-trained models (JMP-S/L; MACE-OFF24 and UMA-S from Appendix~\ref{app:mlip_finetune}); green ring: Atom-JEPA \cite{AtomJEPA_petersen2026}, a concurrent label-free method. The dashed segment marks Orb-v3 (OMol25 pre-trained), whose pre-training hours are unreported (estimated range 5,000--50,000 GPU-hours). SCD hours were measured on NVIDIA A100 GPUs at fp32; JMP, Atom-JEPA, MACE-OFF24 and UMA-S hours are as reported by their authors on different hardware (Table~\ref{tab:mlip_finetuned}). The ET baseline (20.3 meV) is omitted. Results for all ten QM9 targets are in Table~\ref{tab:scd_vs_jmp_qm9}.}
\label{fig:cost_vs_error}
\end{figure}

\begin{figure*}[t]
  \centering
   \includegraphics[width=0.85\linewidth]{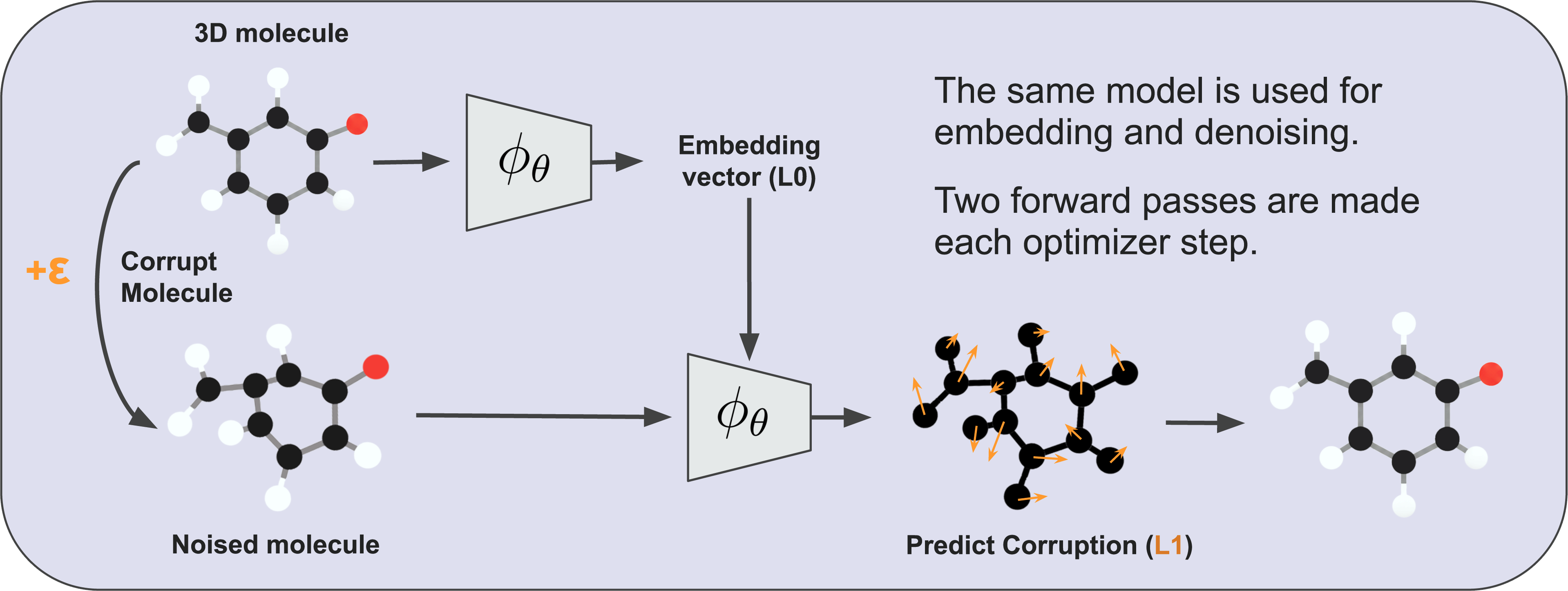}
  \caption{\textbf{Self-Conditioned Denoising Pre-Training.} A first forward pass embeds the clean geometry into a single invariant (L0) embedding vector. This self-embedding conditions a second forward pass (hence self-conditioned), in which the model predicts the per-atom corruption noise ($\boldsymbol{\varepsilon}$) from its equivariant (L1) output. The same model is used for both passes, and both passes occur in each optimizer step.}
  \label{fig:scd}
\end{figure*}

\section{Introduction}

Foundation models have transformed the development pipeline for machine learning in vision and language. The prevailing paradigm pre-trains on large unlabeled datasets for generalizable representations, before subsequent training on smaller, task-specific, labeled data. This success has motivated analogous foundation models in the atomistic sciences, spanning chemistry, structural biology, and materials science \cite{AtomisticfFoundationModels_yuan2025, AI4sci_review_zhang2025artificial}, with the promise of accelerated simulation, structure-to-property prediction, and inverse design.

In chemistry and materials science, the most notable progress toward this vision has come from machine-learned interatomic potentials (MLIPs). Pre-trained MLIPs such as UMA \cite{UMA_wood2025}, MACE \cite{MACE_batatia2025} and Orb \cite{Orbv3_rhodes2025} reach near-DFT accuracy in force-energy prediction through supervised pre-training on large, diverse atomistic datasets. Although often termed `foundation models' (or more appropriately `foundation potentials'), they differ from their language and vision counterparts in that they are pre-trained exclusively by supervised learning on precomputed forces and energies. However, scaling supervised pre-training with DFT labels is computationally expensive; generating forces and energies for the 100M geometries in the OMol25 dataset required 6.6 billion CPU core-hours \cite{OMOL25_levine2025open}.

In other fields, foundation models have depended on self-supervised learning (SSL) objectives that leverage vast, unlabeled datasets. However, unlike images and text, which are often collected without labels, many atomistic datasets are computationally derived (e.g., via DFT) and inherently include force-energy labels. Although force and energy values vary depending on the DFT functional used, prior work has overcome this by using multiple prediction heads and demonstrating clear benefits to supervised pre-training on large, diverse datasets containing labels derived from multiple functionals \cite{UMA_wood2025, JMP_Shoghi2023}. Consequently, supervised pre-training with DFT labels remains the dominant paradigm, outperforming existing SSL methods on simulation and property prediction benchmarks \cite{JMP_Shoghi2023, NoisyNodes_Zaidi2023PVD}. However, the fraction of atomistic data carrying true DFT labels is likely to shrink as more geometries are produced by MLIP-driven simulation, relaxation, and generative models \cite{boltz1_2024, boltz2_2025, Sair_dataset, UMA_wood2025, mattergen_zeni2023, MACE_batatia2025, GeomRep_li2024, Zatom1_2026}. Nevertheless, adapting SSL methods from other fields has remained challenging due to the unique attributes of 3D atomistic data. Atomistic data has no fixed size (samples ranging from 3 to over 3000 atoms), samples may have periodic boundaries, and a model's internal representation must be sensitive to small changes in atomic position to achieve accurate predictions in geometrically sensitive tasks (structural relaxation, simulation, binding affinity prediction, etc.).

Prior work in computer vision suggests that self-supervised pre-training can yield more interpretable and generalizable representations than supervised learning \cite{Mocov3_chen2021, Dino2_oquab2024, DINO3_siméoni2025dinov3}, achieving performance competitive with supervised models even on identical datasets \cite{MAE_he2021, ReLicv2_tomasev2022}. This is because supervised learning only requires a model to represent features of the training data that are relevant to a specific task, becoming invariant to uncorrelated features. In contrast, generative SSL encourages a model to represent the full feature distribution within a training dataset, often producing internal representations with superior transferability. In this work, we demonstrate that this principle holds for 3D atomistic data as well. We do so with Self-Conditioned Denoising (SCD), in which the model's own embedding of a clean geometry conditions the denoising of a corrupted copy of that geometry. Along the way, we provide a systematic evaluation of fine-tuned foundation potentials as property predictors (Appendix~\ref{app:mlip_finetune}), a comparison that has received little attention.

\begin{figure*}[t]
\centering
\includegraphics[width=\textwidth]{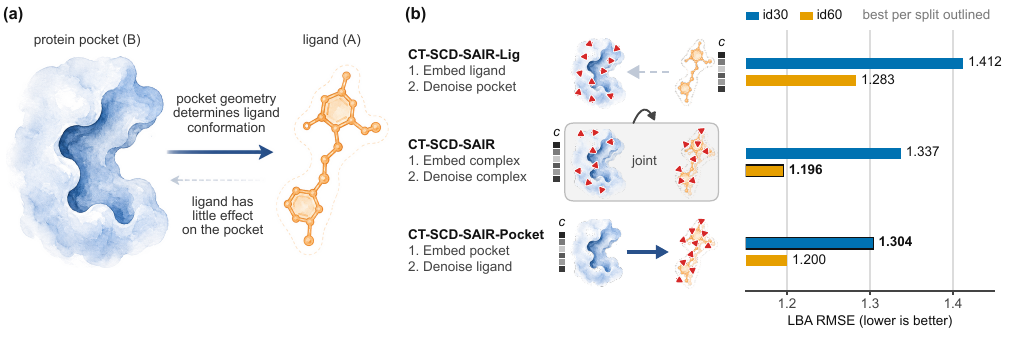}
\caption{\textbf{Component-conditional denoising on pocket-ligand pairs.} \textbf{(a)} Ligand conformation depends strongly on pocket geometry (solid arrow); the reverse dependence is weak (dashed arrow). Here $B$ is the pocket and $A$ the ligand in Equation~\ref{component_conditional_denoising} (Appendix~\ref{PairConditionalDenoising}). \textbf{(b)} Three pre-training objectives on SAIR differ only in the direction of conditioning: embedding the ligand to denoise the pocket runs against the physical dependence, canonical SCD embeds and denoises the complex jointly, and embedding the pocket to denoise the ligand runs with it. Red triangles indicate the denoised component and $\mathbf{c}$ the embedded one. LBA test RMSE on the 30\% and 60\% sequence-identity splits (Table~\ref{tab:lba_full}) tracks the direction of conditioning; the best value in each split is outlined.}
\label{fig:pocket_conditional}
\end{figure*}

\section{Background}
\subsection{Node Denoising}
Early atomistic SSL maximized mutual information between 3D geometries and lower-dimensional views such as 2D graphs or SMILES \cite{GraphMVP_liu2022, 3dinfomax_stärk2022, attrmasking_hu2020}. These methods improved predictions from 2D inputs but underperformed supervised 3D models and fail to embed conformational state information necessary for tasks such as force/energy prediction or docking.

Node denoising emerged as a more general alternative that operates directly on 3D point clouds. Given a structure with $N$ atoms $\mathbf{x} \in \mathbb{R}^{N\times3}$, we sample Gaussian noise $\boldsymbol{\varepsilon} \sim \mathcal{N}(\mathbf{0}, \sigma^{2}\mathbf{I})$ to form a corrupted geometry $\tilde{\mathbf{x}} = \mathbf{x} + \boldsymbol{\varepsilon}$, and train a network $\phi_{\theta}$ to predict the noise on each atom:
\begin{equation}
\label{eq:denoising}
\mathcal{L}_{\text{denoise}} = \mathbb{E}_{\mathbf{x}, \boldsymbol{\varepsilon}} \left[ \left\lVert \phi_{\theta}(\tilde{\mathbf{x}}) - \boldsymbol{\varepsilon} \right\rVert_{2}^{2} \right]
\end{equation}
Unlike contrastive objectives that promote multi-view invariance, noise prediction encourages equivariance to geometric perturbations, making it effective for learning geometric features of local atomic environments. Node denoising has proven useful both as an auxiliary loss \cite{NoisyNodes_Godwin2022, Equiformer1_Liao2022, EquiformerV2_Liao2023} and as a standalone pre-training task \cite{NoisyNodes_Zaidi2023PVD, SE3invDenoise_liu2023, Atomica_Fang2025, EPT_jiao2025, GeoRecon_yan2025}. In computer vision, denoising is likewise an effective self-supervised objective in its own right \cite{DDAE_xiang2023}, and conditioning the denoiser on a compact latent of the clean input, learned end to end, can produce a transferable representation \cite{DiffAE_preechakul2022, SODA_hudson2024}. SCD brings this bottleneck-conditioned design to atomistic data.

\subsection{State-of-the-Art Pre-Training}\label{sec:sota_pretraining}

\begin{table*}[t]
    \centering
    \caption{Summary of Pre-training and Benchmarking Datasets}
    \label{tab:dataset_summary}
    \small
    \setlength{\tabcolsep}{4pt}
    \renewcommand{\arraystretch}{1.15}
    
\begin{tabular}{l l l l l}

    \toprule
         Pre-training Dataset&  Domain&  Unique Structures & Total Geometries& Origin/functional\\
         \midrule
         \textbf{PCQ} (PCQM4Mv2) & Small Molecules &  3.378M &  3.378M & B3LYP/6-31G* DFT\\
         \textbf{GEOM10} & Small Molecules &  304,339&  2.8M& GFN2-xTB, conformers\\
         \textbf{AMP20} (Alex-MP-20)  &  Periodic Materials &  675,204&  675,204& PBEsol DFT\\
         \textbf{SAIR} &  Protein-Ligand &    888k& 4.4M& Boltz-1x, conformers \\
         \textbf{ALL} (AllAtoms) &  Above, mixed & 5.2M & 11.3M & PCQ, GEOM10, AMP20, SAIR \\
         \textbf{OMol25} (4M split) &  Mixed &  ---  &  4M & MD, Non-equilibrium\\
         
        \midrule
        Benchmarking Dataset &  Domain &  Training Set Geometries &  \multicolumn{2}{l}{Task}\\
        \midrule
        \textbf{QM9} & Small Molecules  &  110k &  \multicolumn{2}{l}{properties computed from DFT}\\
        \textbf{Matbench} (mp gap) & Periodic Materials  & 85k &  \multicolumn{2}{l}{band gap energy computed from DFT} \\
        \textbf{LBA} & Protein-Ligand & 4k &  \multicolumn{2}{l}{ligand binding affinity ($-\log(K_d)$) } \\
\bottomrule
    \end{tabular}

\footnotesize PCQ \cite{PubChemQC_Nakata2017}, GEOM \cite{GEOM_axelrod2022}, AMP20 \cite{MaterialsProject, Alexandria_1, Alexandria_2}, SAIR \cite{Sair_dataset}, OMol25 \cite{OMOL25_levine2025open},  Boltz-1x \cite{boltz1_2024}, QM9 \cite{QM9_Ramakrishnan2014}, Matbench \cite{matbench_2020}, LBA \cite{LBA_dataset} 
\end{table*} Despite its generality, standard node denoising has struggled to match supervised force-energy (FE) pre-training, largely because it fails to benefit from non-ground-state (`non-equilibrium') geometries in property prediction \cite{ETOREO_feng2023}. Early interpretations suggested that denoising approximates learning a `near-equilibrium' force field \cite{NoisyNodes_Zaidi2023PVD}, and \citet{NonEqDenoise_wang2023} report gains from denoising non-equilibrium conformers when fine-tuning neural potentials. For high-energy configurations, however, the noise vectors $\boldsymbol{\varepsilon}$ (pointing from corrupted to uncorrupted positions) are unlikely to resemble true forces, and the difference between a corrupted input and a valid higher-energy configuration becomes ambiguous, so denoising non-equilibrium geometries is ill-posed without additional information. One solution is a force encoding \cite{DeNS_liao2024}, but this requires force-energy labels and is no longer SSL.

Others have shown that there is little correlation between noise vectors and true DFT forces even for ground-state geometries \cite{SliDe_Ni2025}; Appendix~\ref{denoising_details} gives our reading of why denoising helps. Variations on node denoising have improved performance at the cost of domain-specific constraints, such as torsional noise in \textit{Frad} \cite{Frad_Ni2024} or pseudo-force targets in \textit{SliDe} \cite{SliDe_Ni2025}; others use specialized masking, block-wise denoising, hybrid contrastive objectives, or scaling to 200M geometries \cite{Atomica_Fang2025, UniCorn_Feng2024, EPT_jiao2025, UniMol_Zhou2023}, but often sacrifice generality and remain limited to ground-state small molecules. Concurrent work instead predicts in latent space (Atom-JEPA; \citealp{AtomJEPA_petersen2026}), but restricts pre-training to near-equilibrium structures. To date, general-purpose SSL objectives have not matched supervised FE pre-training on downstream property prediction \cite{JMP_Shoghi2023, HackNIP_kim2025}.

\section{Method}
\subsection{The Limitations of Node Denoising}

We identify three critical limitations of standard node denoising as a pre-training objective:

\begin{description}\setlength{\itemsep}{2pt}\setlength{\parskip}{0pt}
\item[\textbf{(s1) Locality:}] GNNs often struggle with global representations and long-range correlations, and node denoising fails to produce strong global semantic learning: Gaussian noise on atom positions lacks long-range correlation across space, and small perturbations can be estimated accurately from a small, local context window.
\item[\textbf{(s2) Scalar vs. Vector Embedding Pressure:}] Equivariant GNNs partition node embeddings into channels by rotational symmetry group ($L=0, 1, 2, \dots$). Noise prediction is an equivariant vector ($L=1$; hereafter L1) objective and primarily impacts L1 channels; although gradients reach all parameters, there is little pressure on the invariant scalar embeddings ($L=0$; hereafter L0) to develop good semantic representations, yet downstream property prediction is often roto-translation invariant and relies on the L0 channel.
\item[\textbf{(s3) Ambiguity with Non-Equilibrium Structures:}] As discussed in Section~\ref{sec:sota_pretraining}, without additional context a model cannot distinguish a corrupted input from a valid high-energy geometry, so standard node denoising does not benefit from non-equilibrium data.
\end{description}

\subsection{Self-Conditioned Denoising}
Self-Conditioned Denoising (SCD)\footnote{Distinct from `self-conditioning' in diffusion sampling \cite{AnalogBits_chen2023}, which conditions a denoiser on its own previous estimate of the clean sample with no gradient through it; in SCD the condition is a learned embedding of the clean input, trained end to end.} addresses all three shortcomings by adding a conditional embedding to the denoising objective of Equation~\ref{eq:denoising} (Equation~\ref{scd_obj1}):
\begin{equation}\label{scd_obj1}
\mathbb{E}_{q_{\sigma}(\tilde{\mathbf{x}}, \mathbf{x})}
\left[
  \left\lVert
    \phi_{\theta}(\tilde{\mathbf{x}}|\mathbf{c}) - \boldsymbol{\varepsilon}
  \right\rVert_{2}^{2}
\right], \quad
\mathbf{c} = f_{\eta}(\mathbf{x})
\end{equation}
Here, $\mathbf{c} \in \mathbb{R}^{d}$ is an embedding vector that represents the uncorrupted target geometry. This conditioning gives the model enough information to distinguish a corrupted sample from a valid target conformation (addressing s3). The conditional embedding itself then becomes a training objective that develops a meaningful invariant (L0) representation (addressing s2), which in turn pushes the embedding model $f_{\eta}$ toward a better global, semantic representation (addressing s1). In SCD, the same model serves as both the embedding model $f_{\eta}$ and the noise predictor $\phi_{\theta}$ in each optimizer step (Equation~\ref{SCD_obj2}):
\begin{equation}\label{SCD_obj2}
\begin{aligned}
\mathbb{E}_{q_{\sigma}(\tilde{\mathbf{x}}, \mathbf{x})}
\left[
  \left\lVert
    \phi_{\theta}(\tilde{\mathbf{x}}|\mathbf{c}) - \boldsymbol{\varepsilon}
  \right\rVert_{2}^{2}
\right]\\[1ex]
=
\mathbb{E}_{q_{\sigma}(\tilde{\mathbf{x}}, \mathbf{x})}
\left[
  \left\lVert
    \phi_{\theta}(\tilde{\mathbf{x}}|\phi_{\theta}(\mathbf{x})_{L0})_{L1} - \boldsymbol{\varepsilon}
  \right\rVert_{2}^{2}
\right]\\[2ex]
\mathbf{c} = \phi_{\theta}(\mathbf{x})_{L0} \in \mathbb{R}^d, \quad
\phi_{\theta}(\tilde{\mathbf{x}}|\mathbf{c})_{L1} \in \mathbb{R}^{N\times3}\\
\end{aligned}
\end{equation}
Algorithm~\ref{alg:pretrain_finetune} and Figure~\ref{fig:scd} summarize the implementation. First a forward pass is made on the uncorrupted geometry $\mathbf{x}$. The resulting invariant (L0) embeddings are sum-pooled across nodes and passed through a two-layer MLP embedding head, forming a single embedding per geometry. This information bottleneck encourages a semantic representation: one organized by chemical identity (composition, bonding, and conformation) rather than by extensive quantities such as atom count. The embedding then conditions a second forward pass on the corrupted geometry $\tilde{\mathbf{x}}$, whose L1 equivariant output predicts the per-node noise. The two noise scales in Algorithm~\ref{alg:pretrain_finetune} play distinct roles: $\sigma_{\text{corr}}$ generates $\tilde{\mathbf{x}}$, while a much smaller regularization noise $\sigma_{\text{reg}}$ is applied to the clean input of the embedding pass to reduce overfitting (Appendix~\ref{app:noise_scales}). To preserve unconditional behavior, we drop the conditional embedding for a random 20\% of samples in each batch. After pre-training, downstream fine-tuning and inference require only a single pass without conditioning.

The embedding pass is trained end to end: gradients from the denoising loss flow through $\mathbf{c}$ into the first pass, so the embedding learns whatever representation best supports denoising. In principle, a $d$-dimensional embedding of a small molecule could encode its clean geometry outright and make denoising trivial; however, we do not observe this, because strong regularization (weight decay, $\sigma_{\text{reg}}$ on the embedding pass, and condition dropout) penalizes memorization. Further, because $\mathbf{c}$ is invariant to rotation, translation, and atom ordering, reproducing exact positions and elements from it would require learning a canonical atom ordering -- a harder task than denoising itself.

\begin{algorithm}[t]
\small
\renewcommand{\baselinestretch}{1.3}\selectfont
\caption{SCD Pre-Training and Fine-Tuning Pipeline}
\label{alg:pretrain_finetune}
\begin{algorithmic}[1]
\REQUIRE Dataset $\mathcal{D}$, noise scales $\sigma_{\text{corr}}, \sigma_{\text{reg}}$, model $\phi_{\theta}$
\ENSURE Trained parameters $\theta$

\vspace{0.5em}
\STATE \textbf{Pre-training}
\FOR{each minibatch $\mathbf{x} \sim \mathcal{D}$}
    \STATE $(\mathbf{x}, \tilde{\mathbf{x}}, \boldsymbol{\varepsilon}) \leftarrow \textsc{Corrupt}(\mathbf{x}, \sigma_{\text{corr}}, \sigma_{\text{reg}})$ $\triangleright$ create views
    \STATE $(\mathbf{c}_{L0}, \mathbf{v}_{L1}, \mathbf{y}_{L0}) \leftarrow \phi_{\theta}(\mathbf{x})$ $\triangleright$ self-embedding
    \STATE $(\tilde{\mathbf{c}}_{L0}, \tilde{\mathbf{v}}_{L1}, \tilde{\mathbf{y}}_{L0}) \leftarrow \phi_{\theta}(\tilde{\mathbf{x}}, \mathbf{c}_{L0})$ $\triangleright$ conditional denoising
    \STATE $\mathcal{L}_{\text{pre}} \leftarrow \mathrm{MSE}(\tilde{\mathbf{v}}_{L1}, \boldsymbol{\varepsilon})$
    \STATE $\theta \leftarrow \theta - \lambda \nabla_{\theta} \mathcal{L}_{\text{pre}}$
\ENDFOR

\vspace{0.5em}
\STATE \textbf{Fine-tuning / Inference}
\FOR{each labeled sample $(\mathbf{x}, \mathbf{y})$}
    \STATE $(\mathbf{c}_{L0}, \mathbf{v}_{L1}, \mathbf{y}_{L0}) \leftarrow \phi_{\theta}(\mathbf{x})$
    \STATE $\mathcal{L}_{\text{fine}} \leftarrow \mathrm{loss}(\mathbf{y}_{L0}, \mathbf{y})$
    \STATE $\theta \leftarrow \theta - \lambda \nabla_{\theta} \mathcal{L}_{\text{fine}}$
\ENDFOR
\end{algorithmic}
\end{algorithm}

\subsection{Backbone Architectures}
Following established baselines, we implement SCD using TorchMD-Net (also known as Equivariant Transformer or \textit{ET}) as the primary backbone \cite{TorchMD-Net_thölke2022, Frad_Ni2024, SliDe_Ni2025, UniCorn_Feng2024, NoisyNodes_Zaidi2023PVD}. For a fair comparison, all models we train use an embedding dimension of 256, 8 layers, and 8 heads, consistent with previous SSL studies (see Table~\ref{tab:backbone_hparams}). 

The ET backbone is a lightweight, attention-based GNN without the tensor products and angular embeddings now common in equivariant MLIPs. \textit{Frad} \cite{Frad_Ni2024} and \textit{SliDe} \cite{SliDe_Ni2025} require explicit angular information and use a modified `Geometric Equivariant Transformer' (GET); for a direct comparison we evaluate SCD on both ET and the more expressive GET.

To incorporate conditional inputs into these non-conditional backbones, we replace existing pre-attention layer normalization with Adaptive Layer Normalization (\textit{AdaNorm}), analogous to its use in Diffusion Transformers \cite{DiT_peebles2023} and in representation-conditioned generation \cite{RCG_li2024, GeomRep_li2024}. AdaNorm has a negligible impact on inference speed and memory. In contrast, the GET backbone is approximately $4\times$ slower and has twice as many parameters as the ET backbone. We denote the conditional versions of these two architectures as \textbf{CT} (Conditional TorchMD-Net) and \textbf{CGT} (Conditional Geometric TorchMD-Net), respectively. Implementation details and speed comparisons are in Appendix~\ref{Arch_and_implement_appendix}.

\subsection{Datasets and Tasks}
Atomistic model training has often been siloed to one domain at a time: small molecules, biomolecules, or periodic materials. Although recent work pre-trains on increasingly diverse data \cite{Atomica_Fang2025, ADIT_joshi2025, JMP_Shoghi2023}, multi-domain pre-training remains under-explored. We evaluate models pre-trained on a range of single-domain datasets with varying conformational diversity, and on a combined multi-domain aggregate (Table~\ref{tab:dataset_summary}). All pre-trained models are then benchmarked on three representative domain-specific tasks: 1) QM9 HOMO energy for small molecules \cite{QM9_Ramakrishnan2014}, 2) Matbench band gap for periodic materials \cite{matbench_2020}, and 3) ligand binding affinity (LBA) for biomolecules \cite{LBA_dataset}. Dataset descriptions are in Appendix~\ref{appendix_datasets}.

\section{Results}
\begin{figure*}[t]
\centering
\includegraphics[width=\textwidth]{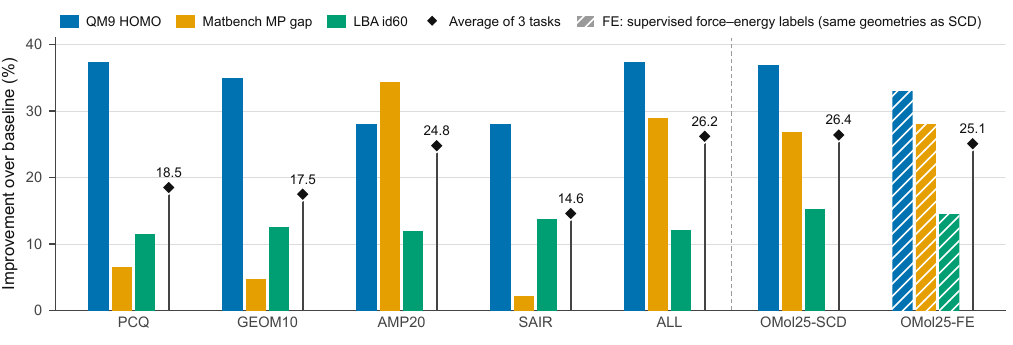}
\caption{\textbf{Every pre-training source helps every task; elemental coverage and conformational diversity decide how much.} Improvement over the non-pre-trained CT baseline on QM9 HOMO (MAE), Matbench MP gap (MAE) and LBA id60 (RMSE) for each SCD pre-training dataset; diamonds mark the three-task average. The two OMol25 models were pre-trained on identical geometries, without (SCD) and with (FE, striped bars) supervised force-energy labels. One run per model; absolute errors are given in Table~\ref{results_summary}.}
\label{fig:data_impact}
\end{figure*}

\begin{table}[t]
\centering
\caption{\textbf{SCD vs. Standard Node Denoising}. Given the same backbone and pre-training dataset (PCQ), SCD outperforms standard denoising (\textit{Coord}). Baseline (ET from scratch) and Coord (node denoising on ET) are the values reported by \citet{TorchMD-Net_thölke2022} and \citet{NoisyNodes_Zaidi2023PVD}; SCD is this work, one run per value. Best results are \textbf{bolded}. All values reported as MAE on the test set.}
\label{tab:scd_vs_coord}
\small
\setlength{\tabcolsep}{5pt} \begin{tabular*}{\columnwidth}{@{\extracolsep{\fill}}l r r r r}
\toprule
\textbf{Method} & \textbf{Baseline} & \textbf{Coord} & \textbf{SCD} & \textbf{\% imp.} \\
Backbone & ET & ET & CT & vs. Coord \\
\midrule
HOMO (meV) & 20.3 & 17.7 & \textbf{12.7} & 28.2\% \\
LUMO (meV) & 17.5 & 14.3 & \textbf{11.5} & 19.6\% \\
Gap (meV)  & 36.1 & 31.8 & \textbf{24.5} & 23.0\% \\
ZPVE (meV) & 1.84 & 1.71 & \textbf{1.18} & 31.0\% \\
$U_0$ (meV) & 6.15 & 6.57 & \textbf{3.58} & 45.5\% \\
$U$ (meV)   & 6.38 & 6.11 & \textbf{3.50} & 42.7\% \\
$H$ (meV)   & 6.16 & 6.45 & \textbf{3.52} & 45.4\% \\
$G$ (meV)   & 7.62 & 6.91 & \textbf{5.29} & 23.4\% \\
$\alpha$ ($a_0^3$) & 0.059 & 0.0517 & \textbf{0.0377} & 27.1\% \\
$C_v$ (cal/mol K) & 0.026 & \textbf{0.020} & 0.021 & $-$5.0\% \\
\bottomrule
\end{tabular*}
\vspace{0.25em}
\end{table} \begin{table}[t]
\centering
\caption{\textbf{SCD vs. Variations on Node Denoising}. All models are pre-trained on PCQ; Frad and SliDe results are taken from the original publications; SCD (CT, CGT) is this work, one run per value. Percent improvement (\% imp.) is calculated between SliDe and SCD pre-training with the most similar backbone, CGT. \textbf{Bold} denotes best; \underline{underline} denotes second best. All values reported as MAE on the test set.}
\label{tab:scd_vs_frad_slide}
\small
\setlength{\tabcolsep}{3pt} \begin{tabular*}{\columnwidth}{@{\extracolsep{\fill}}l r r r r r}
\toprule
\textbf{Method} & \textbf{Frad} & \textbf{SliDe} & \textbf{SCD} & \textbf{SCD} & \textbf{\% imp.}\\
Backbone & GET & GET & CT & CGT & vs. SliDe \\
\midrule
HOMO (meV) & 15.2 & 13.6 & \underline{12.7} & \textbf{9.65} & 29.0\% \\
LUMO (meV) & 13.7 & 12.3 & \underline{11.5} & \textbf{9.05} & 26.4\% \\
Gap (meV)  & 27.8 & 26.2 & \underline{24.5} & \textbf{19.7} & 24.8\% \\
ZPVE (meV) & 1.42 & 1.52 & \textbf{1.18} & \underline{1.22} & 19.7\% \\
$U_0$ (meV) & 5.33 & 4.28 & \textbf{3.58} & \underline{3.96} & 7.5\% \\
$U$ (meV)   & 5.62 & 4.29 & \textbf{3.50} & \underline{4.11} & 4.2\% \\
$H$ (meV)   & 5.55 & 4.26 & \textbf{3.52} & \underline{4.00} & 6.1\% \\
$G$ (meV)   & 6.19 & 5.37 & \textbf{5.29} & \textbf{5.29} & 1.5\% \\
$\alpha$ ($a_0^3$) & \underline{0.0374} & \textbf{0.0366} & 0.0377 & 0.0383 & $-$4.6\% \\
$C_v$ (cal/mol K) & 0.020 & \textbf{0.019} & 0.021 & \textbf{0.019} & 0.0\% \\
\bottomrule
\end{tabular*}
\vspace{0.25em}
\end{table} \label{results}

\subsection{SCD Outperforms Prior Denoising Objectives on Matched Data}\label{sec:matched}
Given the same backbone and pre-training dataset (PCQ), SCD outperforms prior methods of node denoising on QM9 property prediction. Relative to standard node denoising (aka `Coord'; \citealp{NoisyNodes_Zaidi2023PVD}), SCD on the CT backbone lowers error by 19.6--45.5\% on all QM9 targets except heat capacity ($C_v$, $-$5\%) (Table~\ref{tab:scd_vs_coord}). In Table~\ref{tab:scd_vs_frad_slide} we compare SCD against state-of-the-art methods Frad and SliDe. With backbone and pre-training data matched, SCD pre-training on either architecture (CT or CGT) outperforms the published Frad and SliDe results on most QM9 targets, despite minor differences in fine-tuning protocol (Appendix~\ref{appendix_datasets}). Table~\ref{tab:scd_vs_SSL_qm9} in the Appendix provides an additional comparison between our SCD pre-trained models and other methods that used different backbone architectures and/or different pre-training datasets.

\subsection{Labels Are Not Required for Good Representation: SCD vs. Force-Energy Pre-Training}\label{sec:labels_not_required}
\begin{table*}[t]
\centering
\caption{\textbf{SCD Pre-training on Diverse Data Sources}. QM9 HOMO error after SCD pre-training on ground-state and non-equilibrium geometries from varied sources; every configuration improves on the non-pre-trained baseline.}
\label{tab:scd_on_conformers}
\small
\begin{tabular*}{\textwidth}{@{\extracolsep{\fill}}l l r r r}
\toprule
\textbf{Pre-training Dataset} & \textbf{Source / Quality} & \textbf{Unique Structures} & \textbf{Total Geom.} & \textbf{QM9 HOMO} \\ 
& & & & (MAE, meV) \\
\midrule
baseline &   &   &   & 20.3 \\ 
\midrule
\textbf{PCQ} & DFT (B3LYP/6-31G*) & 3,378,606 & 3,378,606 & \textbf{12.7} \\ 
\textbf{PCQ-300k} & DFT (B3LYP/6-31G*) & 300,000 & 300,000 & 13.0 \\ 
\textbf{GEOM10} & xTB (Conformers) & 304,339 & 2,791,929 & 13.2 \\ 
\textbf{GEOM1} & xTB (Lowest Energy) & 304,339 & 304,339 & 13.7  \\ 
\textbf{AMP20} & DFT (PBEsol) & 675,204 & 675,204 & 14.6 \\ 
\textbf{SAIR} & Boltz-1x (Synthetic Conformers) & 888,104 & 4.4M & 14.6 \\ 
\textbf{ALL} & Mixed Aggregate & 5,246,253 & 11.3M & \textbf{12.7} \\ 
\textbf{OMol25}\textsuperscript{1} & Non-equilibrium (MD) & --- & 3,986,754 & 12.8 \\ 
\bottomrule
\end{tabular*}
\vspace{0.25em}
\flushleft\footnotesize \textsuperscript{1} The number of unique molecules in the OMol25 4M split is not reported \cite{OMOL25_levine2025open}.
\end{table*} To compare SCD directly with supervised force-energy (FE) pre-training, we pre-train two CT models on the identical 4M split of OMol25: one by FE prediction using the provided labels, and one by SCD using only the geometries. Figure~\ref{fig:data_impact} summarizes downstream results for each pre-training dataset (absolute errors in Table~\ref{results_summary}, Appendix~\ref{app:summary_tables}). We find that the same model pre-trained using the SCD objective performs roughly as well as FE pre-training on downstream property prediction tasks (MP gap, $-$1.1\%; QM9 HOMO, +3.9\%; LBA, +0.8\%). Each value is a single training run, so we read these margins as parity. The FE baseline shown is the best of several FE pre-training and fine-tuning protocols (see Appendix~\ref{app:fe_pretrain} for details).

\subsection{SCD Benefits From Pre-Training on Any Atomistic Data Source}\label{sec:any_data}
Using all available atomistic data for pre-training remains a bottleneck for atomistic foundation models: prior SSL work has been limited to ground-state small molecules \cite{Frad_Ni2024, SliDe_Ni2025, UniCorn_Feng2024}, and force-energy pre-training across DFT functionals often requires dataset-specific heads or other specialized components \cite{JMP_Shoghi2023, UMA_wood2025}. We find that SCD pre-training benefits from every dataset tested. Table~\ref{tab:scd_on_conformers} shows how datasets from different domains, and of different conformational diversity, affect QM9 HOMO error. SCD benefits from including multiple conformers (GEOM1 vs.\ GEOM10) and non-equilibrium structures (OMol25). We also find that large datasets are not always required for good results: pre-training on a random 10\% subset of PCQ (300k samples) gives only slightly lower performance than the whole dataset (3.4M samples). Overall, the strongest predictor of downstream accuracy is the similarity between the pre-training and fine-tuning datasets. We also find that better-performing models have smoother, less extensive latent spaces; UMAP projections of the learned embeddings and their relation to downstream error are presented in Appendix~\ref{app:umap}.

\subsection{SCD Transfers Across Domains}\label{sec:transfer}
SCD also transfers to materials and proteins. In Table~\ref{tab:mpgap_brief}, our 10M-parameter model pre-trained on 675k periodic materials reaches band gap accuracy within 2\% of JMP-S, a $\sim$30M-parameter model trained on 120M force-energy labeled geometries (108M of them materials). Table~\ref{tab:lba_brief} shows a similar outcome for ligand binding affinity (LBA) on protein-ligand complexes. Here we show how the SCD objective can exploit conditional relationships in atomistic data that are derived from domain knowledge. In the case of protein pocket-ligand pairs, we know that bound ligand conformations are determined by pocket geometry. With `pocket-conditional ligand denoising' (CT-SCD-SAIR-Pocket) we can exploit this fact and provide a task-specific benefit to ligand binding affinity prediction, allowing our pre-trained models to outperform much larger models and achieve state-of-the-art results (Table~\ref{tab:lba_brief}). Conditioning in the opposite direction (`ligand-conditional pocket denoising') harms performance as compared to joint denoising (Figure~\ref{fig:pocket_conditional}; Appendix~\ref{PairConditionalDenoising}).

\subsection{Compute Cost Against Larger Pre-Trained Models}\label{sec:compute}
Figure~\ref{fig:cost_vs_error} places pre-training compute against QM9 HOMO error for SCD, force-energy pre-trained models, the concurrent label-free Atom-JEPA \cite{AtomJEPA_petersen2026}, and the fine-tuned foundation potentials of Appendix~\ref{app:mlip_finetune}. A 10M-parameter CT pre-trained by SCD for 298 GPU-hours (NVIDIA A100, fp32, no compilation) reaches lower error than EquiformerV2 trained from scratch and approaches JMP-S accuracy at one twentieth of its pre-training compute; including fine-tuning on ten QM9 targets, its total cost is also below training EquiformerV2 from scratch (Table~\ref{tab:scd_vs_jmp_qm9}, Appendix~\ref{app:summary_tables}), and it remains competitive with models pre-trained by supervised force-energy prediction on orders of magnitude more data \cite{JMP_Shoghi2023}. GPU-hours for JMP, EquiformerV2 and the foundation potentials are those reported by their authors on different hardware (estimated for Orb-v3). Appendix~\ref{app:mlip_finetune} compares SCD with fine-tuned foundation potentials on all three tasks; full fine-tuning gave the best results in every case, and training only a head on a frozen potential did not reliably beat training the same architecture from scratch (Table~\ref{tab:mlip_ablation}). UMA-S reaches lower HOMO error than any SCD model, at roughly 150$\times$ the pre-training compute and 29$\times$ the parameters; MACE-OFF24 and Orb-v3 (trained on full OMol25) do not.

\begin{table}[h]
\centering
\caption{\textbf{Matbench Band Gap Prediction}. SCD models are competitive with far fewer parameters and unlabeled structures. Within each category best results are \textbf{bolded}; second best are \underline{underlined}.}
\label{tab:mpgap_brief}
\small
\setlength{\tabcolsep}{3pt}
\resizebox{\columnwidth}{!}{\begin{tabular}{l c r r}
\toprule
\textbf{Model} & \textbf{MAE (eV)} & \textbf{Params} & \textbf{Data} \\
\midrule
\textbf{No Pre-training} \\
coGN \cite{CoGN_ruff2023} & \textbf{0.156} & --- & n/a \\ 
CT (Baseline) & \underline{0.186}\textsuperscript{\textdagger} & 10M & n/a \\
JMP-S \cite{JMP_Shoghi2023} & 0.235\textsuperscript{\textdagger} & $\sim$30M & n/a \\ 
JMP-L \cite{JMP_Shoghi2023} & 0.228\textsuperscript{\textdagger} & 235M & n/a \\ 
MODNet \cite{MODNet_DeBreuck2021} & 0.220 & --- & n/a \\ 
\midrule
\textbf{FE Pre-trained} \\
JMP-L \cite{JMP_Shoghi2023} & \textbf{0.091} & 235M & 120M \\ 
UMA-S (1p2), fine-tuned \cite{UMA_wood2025} & \underline{0.097}\textsuperscript{\textdagger} & 291M & $\ge$460M \\
Orb-v3 (inf-mpa), fine-tuned \cite{Orbv3_rhodes2025} & 0.106\textsuperscript{\textdagger} & 24M & $\ge$12M \\
JMP-S \cite{JMP_Shoghi2023} & 0.121 & $\sim$30M & 120M \\
CT-FE-OMol25 & 0.134\textsuperscript{\textdagger} & 10M & 4M \\
HackNIP \cite{HackNIP_kim2025} & 0.150 & $>25$M & $>32$M\textsuperscript{*} \\
\midrule
\textbf{Other SSL Pre-trained} \\
Atom-JEPA \cite{AtomJEPA_petersen2026} & 0.162 & 39M & 1.7M \\
\midrule
\textbf{SCD Pre-trained (Ours)} \\
CT-SCD-AMP20 & \textbf{0.123} & 10M & 675k \\
CT-SCD-ALL & \underline{0.132}\textsuperscript{\textdagger} & 10M & 11.3M \\
CT-SCD-OMol25 & 0.136\textsuperscript{\textdagger} & 10M & 4M \\
\bottomrule
\end{tabular}}
\vspace{0.25em}
\flushleft\footnotesize \textsuperscript{\textdagger} Reported from Fold 0. See full results in Table~\ref{tab:mpgap_full}. \\
\textsuperscript{*} Estimated from Orb-v2 backbone.
\end{table}

\begin{table}[h]
\centering
\caption{\textbf{Ligand binding affinity:} RMSE on the 30\% (id30) and 60\% (id60) sequence-identity splits. Best results are \textbf{bolded}; second best are \underline{underlined}, per split.}
\label{tab:lba_brief}
\small
\setlength{\tabcolsep}{3pt}
\begin{tabular*}{\columnwidth}{@{\extracolsep{\fill}}l r r}
\toprule
\textbf{Model} & \textbf{RMSE: id30 / id60} $\downarrow$ & \textbf{Params} \\
\midrule
\textbf{No Pre-training} \\
ProtNet \cite{ProtNet_wang2023} & 1.463 / 1.343 & --- \\
EPT-Scratch \cite{EPT_jiao2025} & 1.378 / 1.277 & 30M \\
CT (Baseline) & 1.510 / 1.386 & 10M \\
\midrule
\textbf{Other Pre-trained} \\
Frad \cite{Frad_Ni2024} & 1.365 / 1.213 & 14M \\
Uni-Mol \cite{UniMol_Zhou2023} & 1.520 / 1.619 & 47.6M \\
EPT-Protein \cite{EPT_jiao2025} & 1.329 / 1.235 & 30M \\
EPT-Multi \cite{EPT_jiao2025} & 1.322 / 1.227 & 30M \\
ADiT-S \cite{lba-adit} & 1.337 / 1.413 & 12M \\
ADiT-M \cite{lba-adit} & 1.353 / 1.335 & 35M \\
ADiT-L \cite{lba-adit} & \underline{1.308} / 1.246 & 253M \\
\midrule
\textbf{SCD Pre-trained (Ours)} \\
CT-SCD-ALL & 1.372 / 1.218 & 10M \\
CT-SCD-PCQ & 1.332 / 1.226 & 10M \\
CT-SCD-SAIR & 1.337 / \underline{1.196} & 10M \\
CT-SCD-SAIR-Pocket & \textbf{1.304} / 1.200 & 10M \\
CT-SCD-OMol25 & 1.389 / \textbf{1.175} & 10M \\
\bottomrule
\end{tabular*}
\footnotesize Full results provided in Table~\ref{tab:lba_full}.
\end{table}

\section{Discussion}

We introduced Self-Conditioned Denoising, a backbone-agnostic self-supervised objective in which a model's own embedding of a clean geometry conditions the denoising of a corrupted copy. Across eight pre-training datasets and three property-prediction domains, SCD improves every task, matches force-energy pre-training on identical geometries, and lets a 10M-parameter model compete with far larger supervised models at a fraction of the GPU-hours. Four observations follow.

\textbf{All data is useful for pre-training:} Every source tested (DFT ground states, xTB and diffusion-generated conformers, non-equilibrium MD) improved every task, and the benefit crosses domain boundaries: pre-training on periodic materials improves ligand binding affinity prediction by 12\%, and OMol25, a non-periodic dataset spanning 83 elements, improves materials band gap predictions by 26.9\% (Figure~\ref{fig:data_impact}). What transfers is not periodicity but elemental coverage and conformational diversity, with part of the benefit likely from pre-smoothing of the representation space (Figure~\ref{fig:umap}).

\textbf{Pre-training is not necessarily expensive:}
A simple, fast GNN pre-trained by SCD outperforms EquiformerV2 trained from scratch \cite{EquiformerV2_Liao2023} in both accuracy and total compute when addressing multiple tasks (Figure~\ref{fig:cost_vs_error}; Table~\ref{tab:scd_vs_jmp_qm9}). A random 10\% slice of PCQ (300k molecules) captures 96\% of the full-dataset gain on QM9 HOMO, and 675k materials yield band gap accuracy within 2\% of JMP-S \cite{JMP_Shoghi2023}, a 3$\times$ larger model pre-trained on 120M labeled geometries, 108M of them materials. Efficient pre-training was more cost effective than inductive bias through complex architectural elements.

\textbf{Data relevance is key to performance:}
The strongest predictor of success was similarity between pre-training and fine-tuning data in elemental coverage and conformational diversity; for band gap prediction, inorganic pre-training data helped far more than organic-only data. CT-SCD-ALL had the best average across domains but did not surpass single-domain models on in-domain tasks; we attribute this to the limited expressivity of a 10M-parameter model rather than to representation conflicts, though further studies are needed to confirm this. We expect larger models to gain more from multi-domain data.

\textbf{Pre-training a foundation model does not require labels:}
Force-energy pre-training has proven to be an effective and scalable pre-training strategy when DFT labels are readily available \cite{JMP_Shoghi2023, HackNIP_kim2025}. However, computing large numbers of new ground truth (DFT) forces and energies is extremely expensive and time-consuming. Finding effective methods of pre-training with unlabeled, low-fidelity, or cheaply generated atomistic geometries could enable further generalization of atomistic foundation models without requiring a significant expansion of DFT labeled datasets. We provide SCD as a demonstration that this is possible through self-supervised learning. When controlling for architecture and pre-training geometries (labeled vs.\ unlabeled), SCD pre-training matches the performance of supervised force-energy pre-training in downstream property prediction tasks (Section~\ref{sec:labels_not_required}). Further, SCD benefits from pre-training on every data source tested, including diffusion-generated conformers (SAIR). Conditioning is also a design space rather than a fixed recipe. When data arrive as coupled components, embedding one to denoise the other exploits a known physical dependency, and the direction matters: conditioning the ligand on the pocket matches or beats joint denoising and clearly beats the reverse direction (Figure~\ref{fig:pocket_conditional}; Appendix~\ref{PairConditionalDenoising}). The same construction applies to adsorbate--surface, solute--solvent, and defect--host systems. As generative methods in the atomistic sciences continue to advance, we anticipate an increased availability of large `synthetic' datasets for pre-training. 

\textbf{Limitations and outlook:} We use one backbone family (TorchMD-Net and its geometric variant) at 10M--17M parameters, three downstream tasks, and one training run per configuration, so the small margins in Section~\ref{sec:labels_not_required} are not statistically resolved. The Baseline and Coord columns of Table~\ref{tab:scd_vs_coord}, and all Frad, SliDe, JMP, EquiformerV2, UniCorn, EPT and Atom-JEPA numbers, are taken from their publications rather than re-run (Section~\ref{sec:matched}). How SCD scales with model size and multi-domain data remains open, and on force prediction (MD17) SCD offers little advantage over prior methods (Appendix~\ref{app:md17}).

\section{Conclusion}
SCD is a simple change to node denoising: embed the clean geometry, condition the denoiser on that embedding, and train both with one loss. The change removes the ambiguity that kept denoising from using non-equilibrium data, gives the invariant channel a training signal, and yields representations that transfer across small molecules, materials, and protein-ligand complexes. Without a single force or energy label, SCD matches supervised force-energy pre-training on identical geometries and lets a small, fast GNN compete with models orders of magnitude larger and costlier. As MLIP-driven simulation and generative models make unlabeled geometries more plentiful than DFT labels, objectives like SCD offer a route toward utilizing generated data sources for representation learning. Our code and pre-trained models are available at \url{https://github.com/TyJPerez/SelfConditionedDenoisingAtoms}.

\section*{Acknowledgements}
We thank Kaiming He for valuable feedback and helpful discussions on
this work. This research used resources of the National Energy Research Scientific
Computing Center (NERSC), a Department of Energy User Facility using NERSC award
ALCC-ERCAP 0038200.

\section*{Author Contributions}
\label{sec:author_contributions}
T.J.P. conceived this research, developed the core methodology, implemented the framework, and wrote the manuscript.

L.C.L. was responsible for foundation potential fine-tuning and contributed the foundation-potential results in Tables \ref{tab:mlip_finetuned} and \ref{tab:mlip_ablation}.

B.D. provided critical feedback and assisted with manuscript review and editing.

This project was supervised by R.G.B., who provided the necessary resources to complete this work.

\section*{Impact Statement}
This paper aims to advance applications of machine learning in the atomistic sciences. Specifically, we focus on improving the efficacy of self-supervised pre-training, potentially reducing the need for computationally expensive DFT labels. Any societal consequences or impacts that typically relate to the development of atomistic foundation models also apply here. Foundation models and their derivative tools could lower the cost of discovery in many areas of science, including dual use technologies.

\newpage
\appendix
\onecolumn

\section{Fine-Tuning Foundation Potentials for Property Prediction}
\label{app:mlip_finetune}

\subsection{Results}
Pre-trained foundation potentials have demonstrated very strong performance on force and energy prediction tasks. However, there has been relatively little effort to evaluate foundation potentials on other property prediction tasks. Here we evaluate three training methods: fine-tuning with a frozen backbone (only a prediction head is trained), full fine-tuning (pre-trained weights, all parameters trained), and training from scratch (same architecture, random initialization). In Table~\ref{tab:mlip_finetuned} we present the results from fully fine-tuning several popular pre-trained foundation potentials on the same three property prediction tasks used to evaluate our SCD models. We find that larger models perform well, especially UMA-S, but are very costly to pre-train, and relatively expensive to fine-tune fully. The comparative efficacy of our low cost pre-trained models suggests there may be room to improve pre-training efficiency for foundation potentials. Table~\ref{tab:mlip_ablation} presents the results of training the same architectures from scratch, and fine-tuning with a frozen backbone -- a low cost alternative to full parameter tuning. Notably, we find that training a prediction head with a frozen pre-trained backbone does not always outperform training from scratch (without pre-trained weights). Fine-tuning all parameters provided the best results on all three tasks tested (see Table~\ref{tab:mlip_ablation}c).

\subsection{Fine-Tuning Methods}
\label{app:mlip_finetune_method}
In order to fine-tune foundation potentials for property prediction, we replace each model's force-energy output layers with a simple prediction head. The same prediction head architecture is used across all tasks and pre-trained models. The head applies two per-atom linear layers, a pooling operation, then two more linear layers, all of width 256 with SiLU activations in between, except before pooling. The last layer is zero-initialized, so the initial prediction is zero. Sum pooling is used for QM9 HOMO and mean pooling for MP gap and LBA. Per-atom embeddings are taken from the backbone: for MACE, the concatenated scalar blocks of both message-passing layers; for Orb-v3, the final node embeddings; and for UMA, the scalar component of the node embedding, giving feature widths of 256, 256, and 128, respectively. 
All fine-tuning runs use AdamW with a linear warmup (5\% of training length), cosine decay to LR/100, gradient clipping at $\lVert g \rVert_2 \le 10$, and an exponential moving average of weights (decay 0.999) for all tasks; this was not applied to the SCD models. Weight decay is $10^{-4}$ and $10^{-2}$ for full-network and head-only training, respectively. Target labels are normalized (zero mean and unit standard deviation) based on the training split, and loss is computed by mean squared error. We then select the checkpoint with the minimum validation error, stop early after 10\% of the epoch budget without improvement, and evaluate the test split once at that checkpoint. Head architecture, weight decay, schedule, and epoch budget are fixed across models; learning rate is the only tuned hyperparameter, searched over the same four-point grid, except for the LBA seq-id-60 cells, which reuse the learning rate selected on seq-id-30. Total batch size is 128 for QM9 and MP gap, and 8 for LBA; UMA required four GPUs to reach a batch size of 128 at 32 samples each. These training parameters are summarized in Table~\ref{tab:mlip_hparams}. 

\begin{table*}[t]
\centering
\caption{\textbf{Property Prediction with Fine-Tuned Foundation Potentials}. Test set MAE or RMSE on property prediction tasks from three different domains. Model checkpoints were fine-tuned on tasks matching their pre-training domain. Lower is better. \textbf{Bold} denotes best; \underline{underline} denotes second best.}
\label{tab:mlip_finetuned}
\small
\setlength{\tabcolsep}{2.8pt}
\begin{tabular*}{\textwidth}{@{\extracolsep{\fill}}l r r r r r r r}
\toprule
 & & \textbf{Pre-training} & \textbf{Pre-train} & \textbf{QM9} & \textbf{MP gap} & \multicolumn{2}{c}{\textbf{LBA} (RMSE)} \\
\cmidrule(lr){7-8}
\textbf{Model (checkpoint)} & \textbf{Params} & \textbf{geometries} & \textbf{GPU-h} & \textbf{HOMO (meV)} & \textbf{(eV)} & \textbf{id30} & \textbf{id60} \\
\midrule
MACE-OFF24 (med)~\cite{MACEOFF_kovacs2025}  & 1.4M   & 1.16M      & $\approx$350$^{*}$     & 28.5 & --- & --- & --- \\
MACE-MPA-0 (med)~\cite{MACE_batatia2025}  & 9.06M   & $\ge$12M  & $\approx$2{,}600$^{*}$ & --- & 0.156 & 1.463 & 1.207 \\
Orb-v3 (omol)~\cite{Orbv3_rhodes2025}       & 24M  & $\ge$100M & unpublished    & 17.9 & --- & 1.348 & 1.314 \\
Orb-v3 (inf-mpa)~\cite{Orbv3_rhodes2025}    & 24M  & $\ge$12M & unpublished    & --- & \underline{0.106} & --- & --- \\
UMA-S (1p2)~\cite{UMA_wood2025}          & 291M &  $\ge$460M  & $\ge$46{,}080$^{*}$ & \textbf{7.3} & \textbf{0.097} & 1.349 & 1.205 \\
\midrule
CT-SCD-PCQ           & 10M & 3.4M    & 298            & 12.7 & 0.174 & \underline{1.332} & 1.226 \\
CT-SCD-OMol25        & 10M & 4.0M    & 1{,}790          & 12.8 & 0.136 & 1.389 & \textbf{1.175} \\
CT-SCD-AMP20         & 10M & 675k    & 240            & 14.6 & 0.122 & 1.408 & 1.219 \\
CT-SCD-SAIR-Pocket   & 10M & 4.4M    & 390            & --- & --- & \textbf{1.304} & \underline{1.200} \\
CT-SCD-ALL         & 10M & 11.3M  &  2{,}180  & 12.7 & 0.132 & 1.372 & 1.218 \\
CGT-SCD-PCQ          & 17M & 3.4M    & 1{,}308        & \underline{9.65} & --- & --- & --- \\
\bottomrule
\end{tabular*}
\vspace{0.25em}
\flushleft\footnotesize
$^{*}$Pre-training GPU-h are estimated; these figures are not published by the model authors. Hardware differs between rows (A100 for MACE-OFF24, H200 for UMA-S), so the totals are indicative rather than directly comparable.
\end{table*}

\begin{table*}[t]
\centering
\caption{\textbf{Isolating pre-training benefits from model architecture}. 
Test set MAE or RMSE on property prediction tasks under different training methods. Here we present results of training each architecture from scratch (a), with a frozen pre-trained backbone (b), and compare both to full parameter fine-tuning of pre-trained models (c). In (a) and (b), \textbf{bold} denotes best and \underline{underline} second best. Lower is better.}
\label{tab:mlip_ablation}
\small

\begin{subtable}{\linewidth}
\centering
\caption{\textbf{Trained from scratch}: architecture kept, weights randomized.}
\label{tab:mlip_scratch}
\setlength{\tabcolsep}{4pt}
\begin{tabular*}{\textwidth}{@{\extracolsep{\fill}}l r r r r r}
\toprule
 & & \textbf{QM9} & \textbf{MP gap} & \multicolumn{2}{c}{\textbf{LBA} (RMSE)} \\
\cmidrule(lr){5-6}
\textbf{Architecture} & \textbf{Params} & \textbf{HOMO (meV)} & \textbf{(eV)} & \textbf{id30} & \textbf{id60} \\
\midrule
MACE-OFF24 (med)  & 1.4M    & 31.6 & --- & --- & --- \\
MACE-MPA-0 (med)  & 9.06M   & --- & 0.289 & 1.479 & \underline{1.363} \\
Orb-v3 (omol)        & 24M     & \underline{24.3} & --- & \textbf{1.457} & 1.552 \\
Orb-v3 (inf-mpa)     & 24M     & --- & \underline{0.176} & --- & --- \\
UMA-S (1p2)          & 291M    & \textbf{19.5} & \textbf{0.169} & \underline{1.462} & \textbf{1.298} \\
\bottomrule
\end{tabular*}
\end{subtable}

\vspace{1em}

\begin{subtable}{\linewidth}
\centering
\caption{\textbf{Frozen backbone}: published weights fixed, only the head trained.}
\label{tab:mlip_frozen}
\setlength{\tabcolsep}{4pt}
\begin{tabular*}{\textwidth}{@{\extracolsep{\fill}}l r r r r r}
\toprule
 & & \textbf{QM9} & \textbf{MP gap} & \multicolumn{2}{c}{\textbf{LBA} (RMSE)} \\
\cmidrule(lr){5-6}
\textbf{Model (checkpoint)} & \textbf{Params} & \textbf{HOMO (meV)} & \textbf{(eV)} & \textbf{id30} & \textbf{id60} \\
\midrule
MACE-OFF24 (med)  & 1.4M    & \underline{41.1} & --- & --- & --- \\
MACE-MPA-0 (med)  & 9.06M   & --- & 0.160 & 1.520 & 1.442 \\
Orb-v3 (omol)        & 24M     & 50.4 & --- & \textbf{1.457} & \underline{1.403} \\
Orb-v3 (inf-mpa)     & 24M     & --- & \textbf{0.117} & --- & --- \\
UMA-S (1p2)          & 291M    & \textbf{35.0} & \underline{0.133} & \textbf{1.457} & \textbf{1.302} \\
\bottomrule
\end{tabular*}
\end{subtable}

\vspace{1em}

\begin{subtable}{\linewidth}
\centering
\caption{\textbf{Comparing methods}: scratch / frozen / fine-tuned. \textbf{Bold} marks the best method for each model and task.}
\label{tab:mlip_compare}
\setlength{\tabcolsep}{3pt}
\begin{tabular*}{\textwidth}{@{\extracolsep{\fill}}l r r r r}
\toprule
\textbf{Model (checkpoint)} & \textbf{HOMO (meV)} & \textbf{MP gap (eV)} & \textbf{id30} & \textbf{id60} \\
\midrule
MACE-OFF24 (med)  & 31.6\,/\,41.1\,/\,\textbf{28.5} & --- & --- & --- \\
MACE-MPA-0 (med)  & --- & 0.289\,/\,0.160\,/\,\textbf{0.156} & 1.479\,/\,1.520\,/\,\textbf{1.463} & 1.363\,/\,1.442\,/\,\textbf{1.207} \\
Orb-v3 (omol)     & 24.3\,/\,50.4\,/\,\textbf{17.9} & --- & 1.457\,/\,1.457\,/\,\textbf{1.348} & 1.552\,/\,1.403\,/\,\textbf{1.314} \\
Orb-v3 (inf-mpa)  & --- & 0.176\,/\,0.117\,/\,\textbf{0.106} & --- & --- \\
UMA-S (1p2)       & 19.5\,/\,35.0\,/\,\textbf{7.3} & 0.169\,/\,0.133\,/\,\textbf{0.097} & 1.462\,/\,1.457\,/\,\textbf{1.349} & 1.298\,/\,1.302\,/\,\textbf{1.205} \\
\bottomrule
\end{tabular*}
\end{subtable}
\end{table*}

\begin{table*}[t]
\centering
\caption{\textbf{Training hyperparameters for foundation-potential
benchmarking.}}
\label{tab:mlip_hparams}
\small

\begin{subtable}{\linewidth}
\centering
\caption{Settings held constant across every run.}
\label{tab:mlip_hparams_fixed}
\begin{tabular}{@{}l p{0.58\textwidth}@{}}
\toprule
\textbf{Setting} & \textbf{Value} \\
\midrule
Optimizer & AdamW, $\beta_1{=}0.9$, $\beta_2{=}0.999$, $\epsilon{=}10^{-8}$ \\
Learning-rate schedule & 5\% linear warmup, then cosine decay to LR/100 \\
Weight decay & $10^{-4}$ (from scratch, fine-tuned); $10^{-2}$ (frozen) \\
Weight averaging & Exponential moving average, decay 0.999 \\
Head architecture & 2 linear layers $\rightarrow$ pool $\rightarrow$ 2 linear layers; width 256, SiLU except before pooling \\
Head pooling & sum (QM9 HOMO); mean (MP gap, LBA) \\
\midrule
Epoch budget (QM9 / MP gap / LBA) & 400 / 250 / 150 \\
Early-stop patience & 10\% of the budget (40 / 25 / 15) \\
Early-stop minimum change & 0.2 meV / 1.0 meV / 0.01 RMSE \\
Total batch size (QM9 / MP gap / LBA) & 128 / 128 / 8 \\
Precision & fp32 (MACE); bf16 (Orb-v3, UMA) \\
Hardware & NVIDIA A100 (40 GB); UMA uses 4 cards at 32 samples each \\
\bottomrule
\end{tabular}
\end{subtable}

\vspace{1em}

\begin{subtable}{\linewidth}
\centering
\caption{Learning rate selected for each training method, given as scratch / frozen / fine-tuned.}
\label{tab:mlip_hparams_lr}
\setlength{\tabcolsep}{5pt}
\begin{tabular*}{\textwidth}{@{\extracolsep{\fill}}l c c c}
\toprule
\textbf{Task} & \textbf{MACE} & \textbf{Orb-v3} & \textbf{UMA-S} \\
\midrule
QM9 HOMO & $(1 / 1 / 0.3)\times10^{-3}$ & $(0.1 / 1 / 0.3)\times10^{-3}$ & $(1 / 1 / 0.1)\times10^{-3}$ \\
MP gap & $(1 / 1 / 0.3)\times10^{-3}$ & $(0.03 / 1 / 0.03)\times10^{-3}$ & $(1 / 1 / 0.1)\times10^{-3}$ \\
LBA & $(0.1 / 1 / 0.3)\times10^{-3}$ & $(0.1 / 3 / 0.1)\times10^{-3}$ & $(0.1 / 3 / 0.1)\times10^{-3}$ \\
         \bottomrule
\end{tabular*}
\end{subtable}
\end{table*}

\section{Practical Notes on Node Denoising}
\label{denoising_details}

\subsection{Interpreting Noise Scales}\label{app:noise_scales} 
We find that the scale of noise applied to atomistic structures plays a significant role in its utility as a pre-training or regularization tool. Here we formalize two distinct regimes:

\textbf{Regularizing noise} utilizes a small scale with respect to bond length (single bonds range from $\sim$1.0 to 2.0\AA), such as $\sigma \approx 0.005\text{\AA}$, in order to reduce overfitting. At this scale Gaussian noise does not alter single-double bond distinguishability, and is often used as an auxiliary task \cite{NoisyNodes_Godwin2022,Equiformer1_Liao2022, EquiformerV2_Liao2023}. We find that regularizing noise improves downstream results when used in SCD pre-training (on non-corrupted inputs) and performance during some fine-tuning tasks, but not all.

\textbf{Corruption noise} used for pre-training is typically one order of magnitude larger ($\sigma \approx 0.04\text{\AA}$), and often perturbs molecular geometries enough to disrupt single/double/triple bond identities ($\pm$0.1\AA), but without moving atoms far enough to be considered bond breaking. We find that noise scales much larger than this harm pre-training performance.

\subsection{Reinterpreting Denoising for Atomistic Systems}
Many previous works have suggested a relationship between denoising objectives and learning `force fields'; however, we do not find this to be a useful hypothesis for why denoising benefits representation. It is often implied that the `field' learned by noise prediction is beneficial due to its similarity to DFT forces, yet, as noted in Section~\ref{sec:sota_pretraining}, noise vectors on ground-state geometries correlate poorly with true DFT forces. Consistent with this, we find that for our models trained on ground-truth DFT forces, adding an auxiliary node denoising objective harms performance on property prediction, suggesting a low similarity between these targets (see Table~\ref{tab:fe_finetune}).

Instead, we hypothesize that denoising objectives succeed because they encourage the development of smooth, distinguishable local atom embeddings. SCD complements this by introducing a global semantic objective. Through this lens, we attribute the transferability of force-energy pre-training to its demand for high geometric sensitivity and representational complexity, rather than to its explicit connection to physical principles. Force prediction encourages equivariant sensitivity to local perturbations, while energy prediction promotes global representation. Based on this hypothesis, we suggest that self-supervised learning methods should employ reconstruction or de-corruption objectives that challenge models with both local and global objectives similar to force-energy pre-training and SCD.

\section{Backbone Architecture and Implementation}
\label{Arch_and_implement_appendix}

Self-Conditioned Denoising pre-training can be implemented with any architecture designed for diffusion. To modify an existing non-conditional GNN for SCD pre-training, we replace the pre-norm of each message passing layer with an adaptive layer norm (AdaNorm), similar to a diffusion transformer \cite{DiT_peebles2023}. 

\begin{figure}[t]
  \centering
  \includegraphics[width=0.6\linewidth]{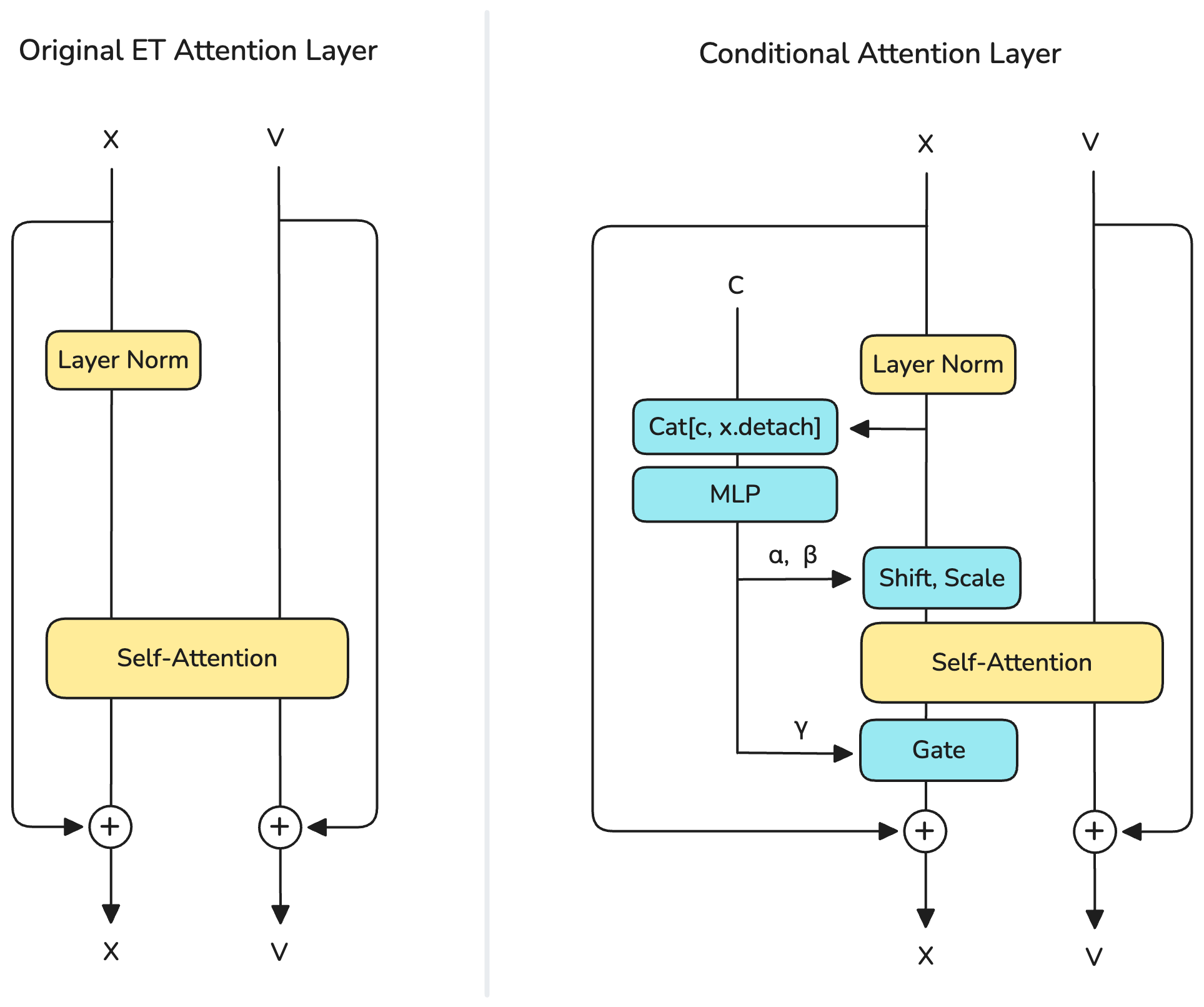}
  \caption{\textbf{Adaptive layer normalization for SCD:} On the left we illustrate a simplified diagram of the original residual layer block in TorchMD-Net (ET). Here, `X' represents invariant (L0) embeddings while `V' represents equivariant vector embeddings (L1), and `C' represents the conditioning embedding. On the right, we show how we modify the original residual block with a conditional scale, shift, and gate of the invariant embeddings. In the gate block the gamma vector is attenuated by a tanh function and multiplied to the output of the ET Attention block. For this work, we use a two-layer MLP for conditioning in each layer.}
  \label{fig:adanorm}
\end{figure}

Figure~\ref{fig:adanorm} diagrams our implementation of AdaNorm used for SCD. Because SCD pre-training requires two forward passes per step, we find that including drop path with a drop probability of 0.1 improves training stability and provides an additional form of regularization. Our drop path is implemented to jointly drop both L0 and L1 embeddings. During pre-training we also freeze atom type/element embeddings. Without freezing element embeddings, we find that element embeddings become vanishingly small, leading to downstream instability. On GitHub \cite{UnstableCoord_Zaidi2023GitHubIssue}, previous authors have noted challenges (gradient explosion) during the downstream fine-tuning of models trained by standard node denoising pre-training (aka `Coord') -- likely a result of small or zeroed out embedding vectors.

Table~\ref{tab:arch_speed_mem} shows a breakdown of parameter count, speed, and memory usage for the architectures used in this work. Each architecture shown uses the same embedding size (256), number of layers (8), and number of attention heads (8). CT and CGT correspond to the modified architectures of ET and GET that incorporate conditional layer norms. In this table, we can see that adding conditional guidance to an architecture has little effect on its speed and memory usage as compared to the addition of edge updates and angle embeddings. CT is overall 4 times faster, almost half the size, and has half the peak memory usage as GET.

\begin{table}[t]
\centering
\caption{\textbf{Architecture Efficiency Comparison}. Benchmarked using QM9 molecules, with batch size 128, averaged over 32 batches, on an NVIDIA RTX A5000 GPU, at fp32 precision with \texttt{torch.compile} off. All models share identical embedding dimensions, layer counts, and attention heads. Parameter counts in this table encompass only backbone architecture, excluding the embedding and prediction heads. All metrics shown are computed from singular forward passes.}
\label{tab:arch_speed_mem}
\small

\begin{subtable}{\linewidth}
\centering
\caption{\textbf{Absolute Metrics}: Latency (lower is better), Throughput (higher is better), and Memory.}
\label{tab:arch_speed_mem_abs}
\setlength{\tabcolsep}{4pt}
\begin{tabular}{l c cc cc cc cc}
\toprule
\textbf{Model} & \textbf{Params} & \multicolumn{2}{c}{\textbf{Latency (ms)} $\downarrow$} & \multicolumn{2}{c}{\textbf{Throughput (samp/s)} $\uparrow$} & \multicolumn{2}{c}{\textbf{Avg Mem (GB)}} & \multicolumn{2}{c}{\textbf{Peak Mem (GB)}} \\
\cmidrule(lr){3-4} \cmidrule(lr){5-6} \cmidrule(lr){7-8} \cmidrule(lr){9-10}
& (M) & Eval & Train & Eval & Train & Eval & Train & Eval & Train \\
\midrule
ET & 6.5 & 65.7 & 69.1 & 1946 & 880 & 0.04 & 0.13 & 1.07 & 8.47 \\
CT & 9.2 & 70.6 & 74.5 & 1812 & 802 & 0.05 & 0.15 & 1.08 & 9.00 \\
GET & 13.4 & 264.9 & 369.5 & 483 & 194 & 0.07 & 0.18 & 1.62 & 20.37 \\
CGT & 16.0 & 262.8 & 379.3 & 487 & 189 & 0.08 & 0.20 & 1.63 & 20.72 \\
\bottomrule
\end{tabular}
\end{subtable}

\vspace{1em}

\begin{subtable}{\linewidth}
\centering
\caption{\textbf{Relative Metrics}: Normalized to ET baseline ($1.0\times$).}
\label{tab:arch_speed_mem_rel}
\setlength{\tabcolsep}{8pt}
\begin{tabular}{l c cc cc}
\toprule
\textbf{Model} & \textbf{Param Ratio} & \multicolumn{2}{c}{\textbf{Rel. Speed} $\uparrow$} & \multicolumn{2}{c}{\textbf{Rel. Avg Mem} $\downarrow$} \\
\cmidrule(lr){3-4} \cmidrule(lr){5-6}
& & Eval & Train & Eval & Train \\
\midrule
ET & $1.00\times$ & $1.00\times$ & $1.00\times$ & $1.00\times$ & $1.00\times$ \\
CT & $1.40\times$ & $0.92\times$ & $0.93\times$ & $1.25\times$ & $1.15\times$ \\
GET & $2.00\times$ & $0.25\times$ & $0.19\times$ & $1.67\times$ & $1.38\times$ \\
CGT & $2.45\times$ & $0.25\times$ & $0.18\times$ & $1.90\times$ & $1.53\times$ \\
\bottomrule
\end{tabular}
\end{subtable}
\end{table} 

\begin{table}[t]
\centering
\caption{Backbone architecture hyperparameters used in this work. CT-small is only used in the MD17 benchmark.}
\label{tab:backbone_hparams}
\begin{tabular}{lccc}
\toprule
\textbf{Hyperparameter} & \textbf{CT} & \textbf{CGT} & \textbf{CT-small} \\
\midrule
Layers                & 8   & 8   & 6   \\
Attention heads       & 8   & 8   & 8   \\
Embedding dimension   & 256 & 256 & 128 \\
Layer pooling         & sum & sum & sum \\
Graph cutoff (\AA)    & 5   & 5   & 5   \\
L0 head pooling       & sum & sum & sum \\
\bottomrule
\end{tabular}
\end{table}

\section{Supplementary Results}

\subsection{Better Models Have Smoother Latent Spaces}\label{app:umap}
\begin{figure*}[t]
  \centering
\includegraphics[width=0.85\linewidth]{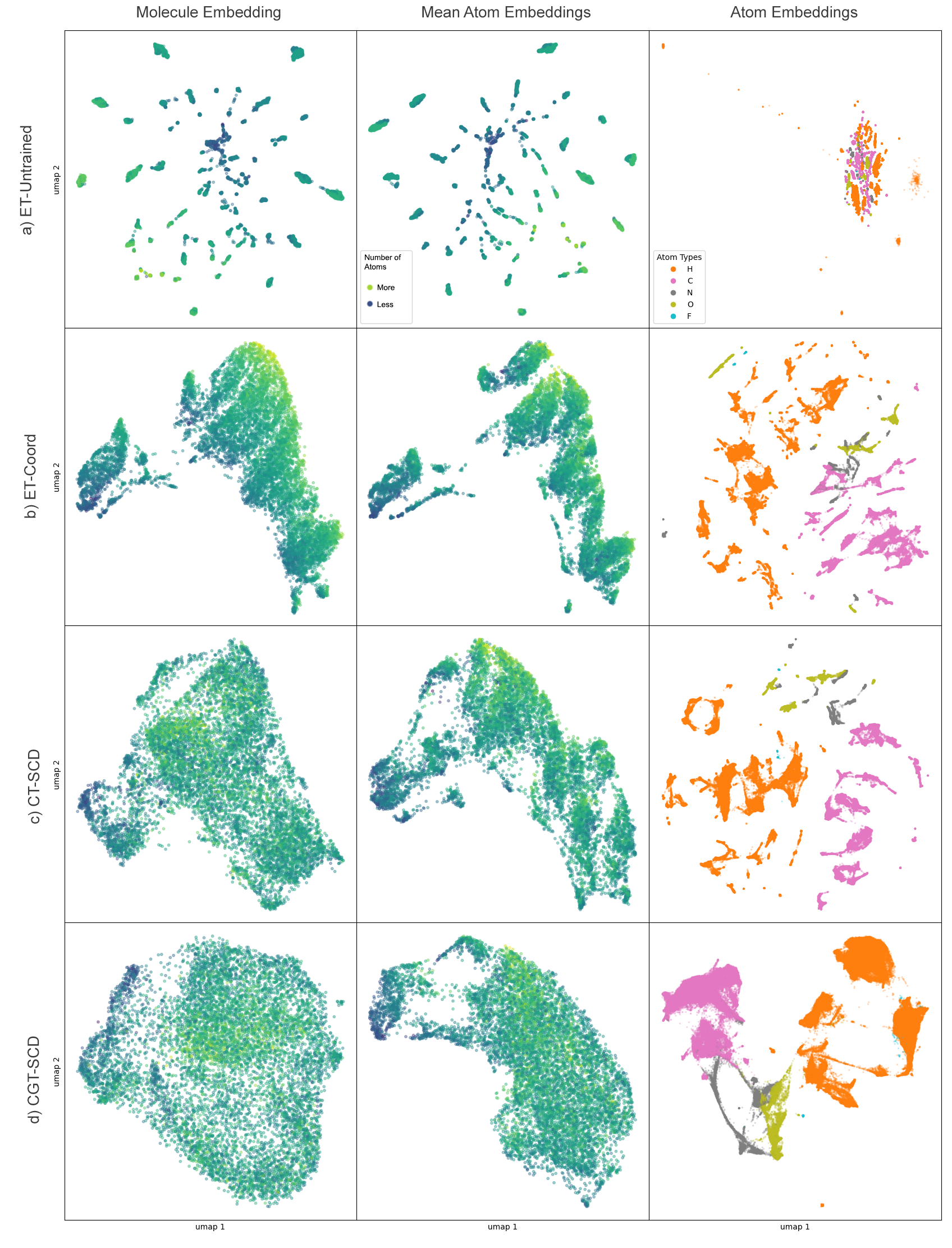}
  \caption{\textbf{Pre-training Effects on the Embedding Space.} UMAP visualizations of embeddings for 10k QM9 molecules, colored by atom count ($N$). \textbf{(a)} Untrained GNNs exhibit a strong bias toward `extensive' representations, clustering molecules primarily by size ($N$). \textbf{(b)} Standard node denoising smooths the latent manifold but embeddings remain highly extensive. \textbf{(c--d)} SCD pre-training yields a smoother, less extensive (more semantic) embedding space. We observe that downstream performance improves as representations become smoother and less extensive (Performance: $a < b < c < d$). `Molecule embeddings' are created by passing sum-pooled atom embeddings through a two-layer MLP `embedding head'. In SCD `Molecule embeddings' are used for self-embeddings. In our Coord model, the molecule `embedding head' remains untrained.}
  \label{fig:umap}
\end{figure*}

Figure~\ref{fig:umap} provides UMAPs for QM9 molecule embeddings from both untrained and pre-trained models, colored by the number of atoms in each sample (N). We observe that an untrained ET model has a strong bias toward highly clustered, extensive representations -- graph-level embeddings that are highly correlated with the number of atoms in each molecule. We find that standard node denoising (Coord) results in a smooth but highly extensive representation space (this was also observed by \citet{GeomRep_li2024}). We believe this is a consequence of features that are too biased toward local node geometry, rather than global semantic representation (non-extensive features). In models pre-trained by SCD, we observe greater smoothing and a reduced correlation with N, suggesting a more semantic representation. Empirically, we find models with a smoother, less extensive, embedding space reach a lower prediction error on QM9 targets (Section~\ref{sec:any_data}).

\subsection{Summary Tables for the Main-Text Figures}\label{app:summary_tables}
Table~\ref{results_summary} lists the absolute errors plotted in Figure~\ref{fig:data_impact}; Table~\ref{tab:scd_vs_jmp_qm9} lists compute and all ten QM9 targets for the models in Figure~\ref{fig:cost_vs_error}.
\begin{table*}[t]
\centering
\caption{\textbf{Comparing the Impact of Different Pre-training Data}. Pre-training on any data source benefits every task, though downstream performance gains are greatest when pre-training on in-domain data. Datasets with high elemental diversity yield the highest average improvements. \textit{CT-FE} models denote supervised force-energy pre-training; \textit{CT-SCD} denotes self-supervised pre-training. Best results are \textbf{bold}; second best are \underline{underlined}. Plotted in Figure~\ref{fig:data_impact}.}
\label{results_summary}
\small
\begin{tabular*}{\textwidth}{@{\extracolsep{\fill}}l r r r r r r r}
\toprule
 & \multicolumn{2}{c}{\textbf{QM9 (Small Mol.)}} & \multicolumn{2}{c}{\textbf{Matbench (Crystals)}} & \multicolumn{2}{c}{\textbf{LBA (Proteins)}} & \\
\cmidrule(lr){2-3} \cmidrule(lr){4-5} \cmidrule(lr){6-7}
\textbf{Model} & \textbf{HOMO} & \textbf{\% imp.} & \textbf{MP Gap} & \textbf{\% imp.} & \textbf{RMSE} & \textbf{\% imp.} & \textbf{Avg \%} \\
(Pre-training Data) & (meV) & & (eV) & & (id60) & & \textbf{imp.} \\
\midrule
ET/CT (Baseline) & 20.3 & 0.0\% & 0.186 & 0.0\% & 1.386 & 0.0\% & 0.0\% \\
\midrule
CT-SCD-PCQ & \textbf{12.7} & 37.4\% & 0.174 & 6.5\% & 1.226 & 11.5\% & 18.5\% \\
CT-SCD-GEOM10 & 13.2 & 35.0\% & 0.177 & 4.8\% & 1.211 & 12.6\% & 17.5\% \\
CT-SCD-AMP20 & 14.6 & 28.1\% & \textbf{0.122} & 34.4\% & 1.219 & 12.0\% & 24.8\% \\
CT-SCD-SAIR & 14.6 & 28.1\% & 0.182 & 2.2\% & 1.196 & 13.7\% & 14.6\% \\
CT-SCD-ALL & \textbf{12.7} & 37.4\% & \underline{0.132} & 29.0\% & 1.218 & 12.1\% & \underline{26.2}\% \\
\midrule
CT-FE-OMol25 & 13.6 & 33.0\% & 0.134 & 28.0\% & \underline{1.187} & 14.4\% & 25.1\% \\
CT-SCD-OMol25 & \underline{12.8} & 36.9\% & 0.136 & 26.9\% & \textbf{1.175} & 15.2\% & \textbf{26.4}\% \\
\bottomrule
\end{tabular*}
\end{table*} \begin{table*}[t]
\centering
\caption{\textbf{Comparing Computational Costs and Performance}. SCD pre-training enables competitive performance at a lower cost than more advanced Clebsch-Gordan tensor product based architectures, and large-scale force-energy (FE) pre-trained models. \textbf{Bold} denotes best; \underline{underline} denotes second best. Plotted in Figure~\ref{fig:cost_vs_error}.
}
\label{tab:scd_vs_jmp_qm9}
\small
\begin{tabular*}{\textwidth}{@{\extracolsep{\fill}}l r r r r r r r}
\toprule
\textbf{Metric} & \textbf{EquiformerV2} & \textbf{Baseline (ET)} & \textbf{CT-SCD} & \textbf{CGT-SCD} & \textbf{Atom-JEPA} & \textbf{JMP-S} & \textbf{JMP-L} \\
\midrule
Model Parameters & 11.2M & 7M & 10M & 17M & 39M & $\sim$30M & 235M \\
Pre-training Structures & 0 & 0 & 3.4M (U) & 3.4M (U) & 19M (U) & 120M (L) & 120M (L) \\
Pre-training GPU-hours & 0 & 0 & \textbf{298} & \underline{1,308} & 2,475 & 5,700 & 34,400 \\
Fine-tuning GPU-h/task & 137 & --- & \textbf{46} & \underline{96} & --- & --- & --- \\
Total hours (10 tasks) & \underline{1,370} & --- & \textbf{758} & 2,268 & $>2,475$ & $>5,700$ & $>34,400$ \\
\midrule
\textbf{QM9 Task (MAE)} & & & & & & & \\
\midrule
HOMO (meV) & 14.0 & 20.3 & 12.7 & \underline{9.65} & 12.2 & 11.1 & \textbf{8.8} \\
LUMO (meV) & 13.0 & 17.5 & 11.5 & \underline{9.05} & 10.9 & 10.8 & \textbf{8.6} \\
Gap (meV) & 29.0 & 36.1 & 24.5 & \underline{19.7} & 24.6 & 23.1 & \textbf{19.1} \\
ZPVE (meV) & 1.47 & 1.84 & 1.18 & 1.22 & 1.06 & \underline{1.0} & \textbf{0.9} \\
$\alpha$ ($a_0^3$) & 0.050 & 0.059 & 0.0377 & 0.0383 & \textbf{0.028} & 0.037 & \underline{0.032} \\
$C_v$ (cal/mol K) & 0.023 & 0.026 & 0.021 & 0.019 & 0.019 & \underline{0.018} & \textbf{0.017} \\
$U_0$ (meV) & 6.17 & 6.15 & 3.58 & 3.96 & 3.38 & \underline{3.3} & \textbf{2.9} \\
$U$ (meV) & 6.49 & 6.38 & 3.50 & 4.11 & 3.38 & \underline{3.3} & \textbf{2.8} \\
$H$ (meV) & 6.22 & 6.16 & 3.52 & 4.00 & \underline{3.28} & 3.3 & \textbf{2.8} \\
$G$ (meV) & 7.57 & 7.62 & 5.29 & 5.29 & 4.99 & \underline{4.5} & \textbf{4.3} \\
\bottomrule
\end{tabular*}
\vspace{0.25em}
\flushleft\footnotesize (U) denotes unlabeled geometries; (L) denotes labeled force-energy structures. SCD GPU-hours were measured on NVIDIA A100 GPUs at fp32 without compilation; EquiformerV2, JMP and Atom-JEPA hours are as reported by their authors on different hardware (Atom-JEPA: NVIDIA H100). The Atom-JEPA parameter count is computed from the authors' released code.
\end{table*} 

\subsection{Ligand Binding Affinity}
\subsubsection{Conditional Denoising of Coupled Components: Pockets and Ligands}
\label{PairConditionalDenoising}
When dataset samples ($\mathbf{x}$) contain interdependent pairs ($\mathbf{x}=\mathbf{A}\cup \mathbf{B}$), such as a protein pocket and a bound ligand, we can alter the self-conditioned denoising objective to instead denoise one component based on an embedding from the other (Equation~\ref{component_conditional_denoising}). In doing so we can exploit known conditional relationships in the dataset during pre-training.

\begin{equation}\label{component_conditional_denoising}
\begin{aligned}
&\mathbb{E}_{q_{\sigma}(\tilde{\mathbf{x}}, \mathbf{x})}
\left[
  \left\lVert
    \phi_{\theta}(\tilde{\mathbf{x}}|\mathbf{c}) - \boldsymbol{\varepsilon}
  \right\rVert_{2}^{2}
\right]
= \mathbb{E}_{q_{\sigma}(\tilde{\mathbf{A}}, \mathbf{B})}
\left[
  \left\lVert
    \phi_{\theta}(\tilde{\mathbf{A}}|\phi_{\theta}(\mathbf{B})_{L0})_{L1} - \boldsymbol{\varepsilon}_A
  \right\rVert_{2}^{2}
\right]\\
\end{aligned}
\end{equation}

In Table~\ref{tab:lba_full} we present results from three variations of pre-training on SAIR, illustrated in Figure~\ref{fig:pocket_conditional}. In CT-SCD-SAIR, we pre-train the model using canonical SCD, embedding and denoising the entire pocket-ligand complexes together. However, in pre-training CT-SCD-SAIR-Lig the ligand structure is embedded and used to conditionally denoise the pocket atoms; and in CT-SCD-SAIR-Pocket we do the reverse, pocket atoms are embedded and used to conditionally denoise the ligand atoms. In doing this we find that CT-SCD-SAIR-Pocket performs the best (specifically on low sequence overlap), CT-SCD-SAIR the second best, and CT-SCD-SAIR-Lig the worst. This follows biochemical intuition, as we would expect ligand conformation to be strongly dependent on pocket geometry, but not the other way around. We hold this as an example of how exploiting known conditional relationships can benefit learned representation during pre-training. We expect component-conditional denoising to transfer to other coupled systems with a known direction of dependence: adsorbates on surfaces, solutes in solvent shells, and dopants or defects in a host lattice, where in each case the environment is the natural conditioning component.
\begin{table*}[t]
\centering
\caption{\textbf{SCD on Proteins}: Ligand Binding Affinity (LBA) predictions over two sequence identity splits (30\% and 60\%). SCD models pre-trained on synthetic or diverse data (e.g., SAIR, OMol25) can outperform larger models pre-trained on PDB-based datasets. Best results within each category are \textbf{bolded}; second best are \underline{underlined}.}
\label{tab:lba_full}
\small
\setlength{\tabcolsep}{2pt} \begin{tabular*}{\textwidth}{@{\extracolsep{\fill}}l l ccc ccc r}
\toprule
\textbf{Method} & \textbf{Model} & \multicolumn{3}{c}{\textbf{Seq. overlap 30\%}} & \multicolumn{3}{c}{\textbf{Seq. overlap 60\%}} & \textbf{Params} \\
\cmidrule(lr){3-5} \cmidrule(lr){6-8}
& & \textbf{RMSE} $\downarrow$ & \textbf{Pear.} $\uparrow$ & \textbf{Spear.} $\uparrow$ & \textbf{RMSE} $\downarrow$ & \textbf{Pear.} $\uparrow$ & \textbf{Spear.} $\uparrow$ & \\

\midrule
\textbf{No Pre-training} 
 & HoloProt-Full Surface \cite{Holoprot_somnath2021multi} & 1.464 & 0.509 & 0.500 & 1.365 & 0.749 & 0.742  & 1.4M\\
 & ProtNet-All-Atom \cite{ProtNet_wang2023}& 1.463 & \underline{0.551} & 0.551 & \underline{1.343} & \underline{0.765} & \underline{0.761}  & --- \\
 & ATOM3D-3DCNN \cite{LBA_dataset} & \underline{1.416} & 0.550 & \underline{0.553} & 1.621 & 0.608 & 0.615  & --- \\
 & ATOM3D-ENN \cite{LBA_dataset}& 1.568 & 0.389 & 0.408 & 1.620 & 0.623 & 0.633  & --- \\
 & ATOM3D-GNN \cite{LBA_dataset}& 1.601 & 0.545 & 0.533 & 1.408 & 0.743 &0.743  & ---\\
 & EPT-Scratch \cite{EPT_jiao2025} & \textbf{1.378} & \textbf{0.604}& \textbf{0.594}& \textbf{1.277}& \textbf{0.787}& \textbf{0.785} &30M\\
 & CT & 1.510& 0.501 & 0.486 & 1.386 & 0.736 & 0.720 &10M\\

\midrule
\textbf{Pre-trained on PCQ} 
& Frad \cite{Frad_Ni2024}& 1.365& 0.599 & 0.577 & \textbf{1.213} & \textbf{0.804} & \textbf{0.801} &14M\\
& CT-SCD-PCQ & \textbf{1.332}& \textbf{0.617}& \textbf{0.600} & 1.226 & 0.800 & 0.794  &10M\\

\midrule
\textbf{Other Pre-trained} 
 & EGNN-PLM \cite{EGNN-PLM_wu2023geometric} & 1.403 & 0.565 & 0.544 & 1.559 & 0.644 & 0.646  & 650M\\
 & Uni-Mol \cite{UniMol_Zhou2023} & 1.520 & 0.558 & 0.540 & 1.619 & 0.645 & 0.653  &47.6M\\
 & ProFSA \cite{profsa_2023} & 1.377 & 0.628 & 0.620 & 1.377 & 0.764 & 0.762  & 47.6M\\
 & EPT-Molecule \cite{EPT_jiao2025} & 1.336& 0.621& 0.602& 1.243& 0.802 & \underline{0.800} &30M\\
 & EPT-Protein \cite{EPT_jiao2025} & 1.329& 0.628& 0.613& \underline{1.235} & \underline{0.804} & \underline{0.800} &30M\\
 & EPT-MultiDomain \cite{EPT_jiao2025}  & \underline{1.322} & \underline{0.644} & \underline{0.630} & 
 \textbf{1.227} & \textbf{0.811} & \textbf{0.803}  &30M\\
& ADiT-S \cite{lba-adit} & 1.337 & 0.626 & 0.618 & 1.413 & 0.740 & 0.740   & 12M \\
& ADiT-M \cite{lba-adit} & 1.353 & 0.622 & \underline{0.630} & 1.335  & 0.764 & 0.752  & 35M \\
& ADiT-L \cite{lba-adit} & \textbf{1.308} & \textbf{0.645} & \textbf{0.647} & 1.246  & 0.767 & 0.765  & 253M\\

\midrule
\textbf{SCD Pre-trained} 
 & CT-SCD-AMP20 & 1.408 & 0.584 & 0.533 & 1.219 & 0.803 & 0.800 & 10M \\
 & CT-SCD-GEOM10 & 1.392 & 0.606 & 0.554 & 1.211 & 0.806 & 0.801 & 10M \\
 & CT-SCD-ALL & 1.372 & 0.594 & 0.578 & 1.218 & 0.802 & 0.798 & 10M \\
 & CT-SCD-SAIR-Lig & 1.412 & 0.566 & 0.547 & 1.283 & 0.779 & 0.770 & 10M \\
 & CT-SCD-SAIR & \underline{1.337} & \underline{0.617} & \underline{0.599} & \underline{1.196} & \underline{0.810} & 0.802 & 10M \\
 & CT-SCD-SAIR-Pocket & \textbf{1.304} & \textbf{0.640} & \textbf{0.624} & 1.200 & 0.809 & \underline{0.806} & 10M \\ 
 & CT-SCD-OMol25 & 1.389 & 0.586 & 0.571 & \textbf{1.175} & \textbf{0.817} & \textbf{0.816} & 10M \\

\midrule
\textbf{FE Pre-trained} 
& CT-FE-OMol25 & 1.391 & 0.575 & 0.564 & 1.187 & 0.814 & 0.811 & 10M \\
\bottomrule
\end{tabular*}
\end{table*} 
\begin{table*}[t]
\centering
\caption{\textbf{SCD on Periodic Materials}: Matbench band gap prediction performance for SCD and supervised force-energy (FE) pre-trained models. Best results within each category are \textbf{bolded}; second best are \underline{underlined}. SCD models achieve competitive performance while utilizing significantly fewer parameters and unlabeled structures.}
\label{tab:mpgap_full}
\small
\begin{tabular*}{\textwidth}{@{\extracolsep{\fill}}l l c c r r}
\toprule
\textbf{Category} & \textbf{Model} & \multicolumn{2}{c}{\textbf{MP Gap (eV)}} & \textbf{\# Params} & \textbf{Pre-training} \\
\cmidrule(lr){3-4}
& & \textbf{Fold 0} & \textbf{Mean (0--4)} & & \textbf{Data Size} \\
\midrule
No Pre-training & MODNet \cite{MODNet_DeBreuck2021} & 0.215 & 0.220 & --- & n/a \\ 
 & coGN \cite{CoGN_ruff2023} & \textbf{0.153} & \textbf{0.156} & --- & n/a \\ 
 & JMP-S \cite{JMP_Shoghi2023} & 0.235 & --- & $\sim$30M &n/a \\ 
 & JMP-L \cite{JMP_Shoghi2023} & 0.228 & --- & 235M & n/a \\ 
 & CT (Baseline) & \underline{0.186} & --- & 10M & n/a \\
\midrule
FE Pre-trained & JMP-L \cite{JMP_Shoghi2023} & \textbf{0.089} & \textbf{0.091} & 235M & 120M \\ 
 & JMP-S \cite{JMP_Shoghi2023} & \underline{0.119} & \underline{0.121} & $\sim$30M & 120M \\
 & HackNIP (Orb-v2 + MODNet) \cite{HackNIP_kim2025} & --- & 0.150 & $>25.2$M & $>32$M\textsuperscript{*} \\
 & CT-FE-OMol25 & 0.134 & --- & 10M & 4M \\

\midrule
Other SSL & Crystal-Twins \cite{xtal_twins} & --- &  \underline{0.264} & --- & 428k \\
 & Atom-JEPA \cite{AtomJEPA_petersen2026} & --- & \textbf{0.162} & 39M & 1.7M \\

\midrule
SCD Pre-trained & CT-SCD-AMP20 & \textbf{0.122} & \textbf{0.123} & 10M & 675k \\
 & CT-SCD-ALL & \underline{0.132} & --- & 10M & 11.3M \\
 & CT-SCD-PCQ & 0.174 & --- & 10M & 3.4M \\
 & CT-SCD-GEOM10 & 0.177 & --- & 10M & 2.8M \\
 & CT-SCD-SAIR & 0.182 & --- & 10M & 4.4M \\
 & CT-SCD-OMol25 & 0.136 & --- & 10M & 4M \\
\bottomrule
\end{tabular*}
\vspace{0.25em}
\flushleft\footnotesize \textsuperscript{*}Orb-v2 data size estimated from the sum of force-energy fine-tuning datasets: MPTraj (1.58M) and Alexandria (30.5M) \cite{HackNIP_kim2025}.
\end{table*} 
\subsection{Force-Energy Pre-Training and Fine-Tuning for Property Prediction}\label{app:fe_pretrain}

In this work, we conduct all force-energy pre-training using the OMol25-4M dataset and a CT backbone. In Table~\ref{tab:fe_pretrain}, we compare models pre-trained via either direct (`forward') or gradient-based (`backward' aka `conservative') force prediction. Unsurprisingly, the simple CT backbone does not yield state-of-the-art force-energy predictions; however, we find that SCD pre-training confers a modest benefit even when restricted to the same training geometries as the labeled task. Additionally, we find that a single SCD step (comprising a double forward pass) is roughly half the speed of a direct force training step (a single forward pass), but $1.4\times$ faster than a gradient-based (forward then `backward') force training update step. Table~\ref{tab:fe_finetune} presents benchmarking results for these models on two representative QM9 tasks: HOMO and $U_0$ (internal energy at 0\,K). We observe that pre-training via direct force prediction benefits downstream property prediction more than gradient-based force pre-training, but it does not outperform SCD pre-training. In main text tables, we report only the best model -- pre-trained exclusively on direct force prediction -- as `CT-FE-OMol25'.

\begin{table}[h]
\centering
\caption{\textbf{OMol25 4M Force-Energy Pre-training}. Comparison of supervised force-energy training baselines (FE) against fine-tuning from SCD pre-trained weights (SCD-FE). We compare direct force prediction (Forward) against gradient-based force calculation (Backward). CT-SCD-OMol25 is pre-trained by SCD using the same OMol25 4M geometries.}
\label{tab:fe_pretrain}
\small
\setlength{\tabcolsep}{5pt}
\begin{tabular}{l l l c cl}
\toprule
\textbf{Model Name} & \textbf{Initialization} & \textbf{Force Method} & \textbf{Val Force MAE} & \textbf{Val Energy MAE}  & \textbf{num training steps}\\
& & & (meV/\AA) & (meV)  &\\
\midrule
FE-Fwd & From Scratch & Forward & 44.8 & 648  &800k\\
FE-Fwd-long & From Scratch & Forward & 42.6 & 606  &1600k\\
FE-Bwd & From Scratch & Backward & 50.0 & 618  &800k\\

\midrule
SCD-FE-Fwd & CT-SCD-OMol25 & Forward & 42.9 \small(+4.2\%) & 617 \small(+4.8\%)  &800k (+800k SCD)\\
SCD-FE-Bwd & CT-SCD-OMol25 & Backward & 41.4 \small(+17.2\%) & 665 \small($-$7.6\%) &800k (+800k SCD)\\
\bottomrule
\end{tabular}
\end{table}

\begin{table}[h]
\centering
\caption{\textbf{Fine-tuning Force-Energy Pre-trained Models}. Force-energy pre-trained models performed best when the energy head was reset and regularization noise was turned off during fine-tuning. Forward (non-conservative) force pre-training conveyed the best average results on downstream property prediction and is reported as `CT-FE-OMol25' elsewhere.
}
\label{tab:fe_finetune}
\small
\setlength{\tabcolsep}{5pt}
\begin{tabular}{l l c cl}
\toprule
\textbf{Model} & \textbf{Fine-tuning Protocol} & \textbf{HOMO} & \textbf{$U_0$}  &\textbf{Avg \% imp.} \\
& & (meV) & (meV)  &over baseline\\
\midrule
 baseline& & 20.3& 6.15&---\\
\midrule
FE-Fwd & Standard & 15.9 & 12.9  &$-$44.0\%\\
FE-Fwd & Reset Head, $\sigma_{\text{reg}}=0.005$ & 15.0 & 10.2  &$-$19.9\%\\
FE-Fwd & Reset Head, $\sigma_{\text{reg}}=0$ & 13.6 & 3.40  &38.9\%\\
FE-Fwd-long & Reset Head, $\sigma_{\text{reg}}=0$ & 13.7 & 3.47  &38.0\%\\
FE-Bwd & Reset Head, $\sigma_{\text{reg}}=0$ & 14.5 & 3.36  &37.0\%\\
\midrule
CT-SCD-OMol25 & Standard & 12.8 & 3.54  & 39.7\%\\
\bottomrule
\end{tabular}
\end{table}

\begin{table}[h]
\centering
\caption{SCD vs. other SSL Methods using different backbones and pre-training data. Models CT and CGT below are both pre-trained on the PCQ dataset. Best results are \textbf{bolded}; second best are \underline{underlined}.}
\label{tab:scd_vs_SSL_qm9}
\small
\setlength{\tabcolsep}{3pt}
\begin{tabular}{l c c c c c}
\toprule
\textbf{QM9 Task} & \textbf{CT} & \textbf{CGT} & \textbf{EPT-10} & \textbf{UniCorn} & \textbf{Atom-JEPA} \\
 & (Ours) & (Ours) & \cite{EPT_jiao2025} & \cite{UniCorn_Feng2024} & \cite{AtomJEPA_petersen2026} \\
\midrule
Params & 10M & 17M & 30M & --- & 39M \\
Pre-train Data & 3.4M & 3.4M & 5.9M & 15M & 19M \\
\midrule
HOMO (meV) & 12.7 & \textbf{9.65} & 15.2 & 13.0 & \underline{12.2} \\
LUMO (meV) & 11.5 & \textbf{9.05} & 13.6 & 11.9 & \underline{10.9} \\
Gap (meV) & \underline{24.5} & \textbf{19.7} & 29.0 & 24.9 & 24.6 \\
ZPVE (meV) & 1.18 & 1.22 & \underline{1.11} & 1.4 & \textbf{1.06} \\
$U_0$ (meV) & \underline{3.58} & 3.96 & 5.44 & 3.99 & \textbf{3.38} \\
$U$ (meV) & \underline{3.50} & 4.11 & 5.54 & 3.95 & \textbf{3.38} \\
$H$ (meV) & \underline{3.52} & 4.00 & 5.42 & 3.94 & \textbf{3.28} \\
$G$ (meV) & 5.29 & 5.29 & 6.37 & \underline{5.09} & \textbf{4.99} \\
$\alpha$ ($a_0^3$) & 0.0377 & 0.0383 & 0.045 & \underline{0.036} & \textbf{0.028} \\
$C_v$ (cal/mol K) & 0.021 & \textbf{0.019} & 0.020 & \textbf{0.019} & \textbf{0.019} \\
\bottomrule
\end{tabular}
\vspace{0.25em}
\flushleft\footnotesize The pre-training datasets for EPT-10 and UniCorn include the 3.4M PCQ structures used by SCD.
\end{table}
 
\subsection{MD17: Small Molecule Forces}\label{app:md17}

Consistent with prior work, we benchmark a smaller SCD model (6 layers, dim 128) on MD17 \cite{md17_chmiela2017machine}. The models presented in Table~\ref{tab:md17_scd} were pre-trained on the PCQ dataset and subsequently fine-tuned on MD17. To match the fine-tuning methods described by Frad \cite{Frad_Ni2024}, we also train a model using a two pass `force-denoise' scheme, in which the first pass is used to predict forces and energies and the second forward pass is used for an auxiliary denoising task. We observe that SCD pre-training with the CT backbone yielded results similar to those of Frad and SliDe, while achieving $4\times$ faster inference speeds. However, none of these methods achieve particularly strong performance on this benchmark. We hypothesize that this stems from the pre-training dataset containing exclusively ground-state geometries; consequently, the non-equilibrium structures in MD17 are likely out-of-distribution. Lacking non-equilibrium structures during pre-training, the models fail to capture features necessary for non-equilibrium prediction tasks. This aligns with our findings on other benchmarks, where similarity between pre-training and fine-tuning data distributions is a strong predictor of performance. We do not report results for the GET/CGT variant due to persistent segmentation faults when computing gradient-based forces.

\begin{table}[t]
\centering
\caption{SCD achieves comparable performance on MD17 force-energy prediction to prior methods while using a smaller and faster architecture. Best results are in \textbf{bold}, second best are \underline{underlined}.}
\label{tab:md17_scd}
\renewcommand{\arraystretch}{1.15}
\setlength{\tabcolsep}{6pt}
\resizebox{\linewidth}{!}{\begin{tabular}{l|cccc!{\vrule width 1.2pt}cc}
\toprule
\multicolumn{1}{l|}{\textbf{Method}}              
& \textbf{Baseline} 
& \textbf{Coord} 
& \textbf{SCD} 
& \textbf{SCD (force-denoise)} 
& \textbf{Frad (force-denoise)} 
& \textbf{SliDe} \\

\multicolumn{1}{l|}{\textbf{Architecture}}        
& ET-small 
& ET-small 
& CT-small 
& CT-small 
& GET-small 
& GET-small \\

\multicolumn{1}{l|}{\textbf{Pre-training dataset}} 
& none 
& PCQ 
& PCQ 
& PCQ 
& PCQ 
& PCQ \\
\midrule
\multicolumn{7}{l}{\textbf{Forces MAE (kcal/mol/\AA)}}\\
\midrule
Aspirin         & 0.253 & 0.211 & 0.210 & \underline{0.200} & 0.209 & \textbf{0.174} \\
Benzene        & 0.196 & \textbf{0.169} & 0.171 & 0.175 & 0.199 & \textbf{0.169} \\
Ethanol        & 0.109 & 0.096 & 0.120 & 0.116 & \underline{0.091} & \textbf{0.088} \\
Malonaldehyde  & 0.169 & \textbf{0.139} & 0.186 & 0.173 & \underline{0.142} & 0.154 \\
Naphthalene    & 0.061 & 0.053 & \underline{0.041} & \textbf{0.040} & 0.053 & 0.048 \\
Salicylic acid & 0.129 & 0.109 & \underline{0.101} & \textbf{0.099} & 0.108 & \underline{0.101} \\
Toluene        & 0.067 & 0.058 & \underline{0.053} & \textbf{0.051} & 0.054 & 0.054 \\
Uracil         & 0.095 & \underline{0.074} & 0.077 & \textbf{0.072} & 0.076 & 0.083 \\
\midrule[1.2pt]
\multicolumn{7}{l}{\textbf{Energy MAE (kcal/mol)}}\\
\midrule
Aspirin         & 0.123 & -- & 0.0742 & \textbf{0.0705} & -- & -- \\
Benzene         & 0.058 & -- & 0.0274 & \textbf{0.0211} & -- & -- \\
Ethanol         & 0.052 & -- & 0.0286 & \textbf{0.0280} & -- & -- \\
Malonaldehyde   & 0.077 & -- & 0.0432 & \textbf{0.0417} & -- & -- \\
Naphthalene     & 0.085 & -- & 0.0163 & \textbf{0.0156} & -- & -- \\
Salicylic acid  & 0.093 & -- & \textbf{0.0286} & 0.0290 & -- & -- \\
Toluene         & 0.074 & -- & \textbf{0.0181} & 0.0184 & -- & -- \\
Uracil          & 0.090 & -- & \textbf{0.0190} & 0.0193 & -- & -- \\
\bottomrule
\end{tabular}
}
\vspace{0.5em}
\flushleft\footnotesize CGT results on MD17 are omitted due to an unresolved segmentation fault during backpropagation.
\end{table} 

\section{Datasets}
\label{appendix_datasets}

\subsection{Pre-Training Datasets}

\textbf{PCQ}: The PCQM4Mv2 dataset is a curated subsample of the PubChemQC project containing ground-state geometries for 3.378M organic small molecules obtained using B3LYP/6-31G* DFT \cite{PubChemQC_Nakata2017}. This dataset is commonly used for demonstrating self-supervised learning methods, either on its own or as part of a composite dataset. Throughout this work we use `PCQ' to indicate models trained only on this dataset.

\textbf{GEOM10}: The original GEOM dataset contains 37M conformers from $\sim$450k unique molecules generated using GFN2-xTB, a semi-empirical extended tight-binding method \cite{GEOM_axelrod2022}. GEOM is split into three categories: 1) GEOM-drugs, containing conformers of drug like small molecules, 2) GEOM-QM9, conformers of molecules from the QM9 dataset, and 3) GEOM-MoleculeNet, containing conformers from a subsample of molecules in the MoleculeNet benchmark. To avoid overlap with benchmarking datasets we only utilize structures from GEOM-drugs, containing 304k unique molecules, and curate two subsamples: GEOM10, containing up to 10 of the lowest energy conformers from each molecule (2.8M geometries total), and GEOM1 containing only the lowest energy (ground-state) conformer from each molecule (304k geometries).

\textbf{SAIR}: The Structurally Augmented IC50 Repository (SAIR) dataset contains a large number of bound protein-ligand structures generated using the Boltz-1x model \cite{boltz1_2024}. Five conformers are provided for each structure. Although the authors of SAIR report 1,048,857 unique binding pairs totaling 5.2M conformers, the copy we obtained from Hugging Face (\url{https://huggingface.co/datasets/SandboxAQ/SAIR}) contains only 888,104 unique binding pairs and 4.4M conformers. 

\textbf{Alex-MP-20}: The Alex-MP-20 dataset comprises 675,204 periodic materials structures with up to 20 atoms \cite{mattergen_zeni2023} derived from both the Materials Project \cite{MaterialsProject}, and the Alexandria materials database \cite{Alexandria_1, Alexandria_2}. Throughout this work, we refer to Alex-MP-20 as `AMP20'. When SCD pre-training on AMP20, we apply a unit cell repeat augmentation to simulate larger crystals. Each sample is randomly tiled in one direction with a probability ($p_{\text{cell repeat}}$) of 50\%, for up to two repeats. This dataset is also available on Hugging Face (\url{https://huggingface.co/datasets/OMatG/Alex-MP-20}).

\textbf{AllAtoms (ALL):} We also pre-train on a combined dataset that includes the entirety of PCQ, GEOM10, SAIR, and AMP20, totaling 5.2M unique structures with 11.3M conformations.

\textbf{OMol25 4M split}: The Open Molecules 2025 (OMol25) dataset is a large-scale repository of over 100 million quantum chemistry calculations covering 83 million unique molecular systems \cite{OMOL25_levine2025open}. While the full dataset spans the first 83 elements (H--Bi) and systems up to 350 atoms, we utilize the designated `4M split', a randomly sampled subset of approximately 4 million geometries. All labels (forces and energies) were computed at the $\omega$B97M-V/def2-TZVPD level of theory. The 4M subset preserves the broad chemical diversity of the full dataset, containing equilibrium and non-equilibrium structures from four primary domains: small organic molecules, biomolecules (proteins, DNA/RNA from PDB/BioLiP2), metal complexes (transition metals, ligands), and electrolytes. This dataset is available on Hugging Face (\url{https://huggingface.co/facebook/OMol25}).

\subsection{Benchmarking Datasets and Fine-Tuning Hyperparameters}

\textbf{QM9}: The QM9 dataset is a commonly used benchmark for small molecule property prediction on 3D geometries \cite{QM9_Ramakrishnan2014}. QM9 comprises a set of small organic molecules containing up to nine heavy atoms (C, N, O, F), or up to 29 atoms total when including hydrogen. All 3D geometries from QM9 are computed using the B3LYP/6-31G(2df, p) DFT functional and are labeled with a set of 15 quantum mechanical properties including HOMO-LUMO gap, atomization energies, specific heat, polarizability and more. The original dataset contains 134k molecules; however, it is now standard practice to use a cleaned version containing 130,831. We obtain our copy from the PyTorch Geometric library. QM9 is typically split randomly: 110,000 for training, 10,000 for validation, and the remaining 10,831 for the test set. We use this convention in our work; however, we do not use the validation set for early stopping. We follow the practices described in TorchMD-Net \cite{TorchMD-Net_thölke2022} and subtract provided reference energies from atomization energy targets. All other targets are normalized to a standard Gaussian for prediction.

\textbf{LBA}: The Ligand Binding Affinity (LBA) dataset comprises 4,463  co-crystallized 3D structures of  protein-ligand complexes derived from PDBbind. Binding affinity targets are experimentally derived and provided as negative log of either inhibition or dissociation constant ($-\log(K)$). Two splits are provided based on sequence overlap between proteins in the training set and test set; in `id60' no protein in the training set has more than a 60\% sequence overlap with any protein in the test set. Similarly, in `id30' train-test sequence overlap is limited to 30\%. Following ATOM3D preprocessing \cite{LBA_dataset}, predictions are made using extracted pocket-ligand complexes.

\textbf{Matbench}: The Matbench properties dataset \cite{matbench_2020} provides 13 different tasks (regression and classification) for periodic materials structures to benchmark property prediction models. Across these tasks, training set size ranges from 312 to 132k samples, each of which is split into 5 folds with predefined test splits. In this work we explore only one regression task, `mp\_gap' or band gap energy prediction, as a proof of concept. We choose band gap energy because of its economic relevance. The `mp\_gap' task provides 106k periodic crystal structures in total, resulting in training set sizes of 85k and test set sizes of 21k. We report results for fold 0 with all models tested. For our best model we also report the average of all folds.

\textbf{MD17}: The MD17 \cite{md17_chmiela2017machine} dataset comprises a collection of ab initio molecular dynamics (AIMD) trajectories for eight small organic molecules with DFT forces and energies, commonly used to evaluate MLIPs. Following past convention we use a random split of 950 samples for training, 50 for validation (however we do not employ early stopping), and the remainder as the test set. The total number of geometries provided for each trajectory varies from 150k to 1M. All force predictions provided are derived from energy gradients (a backwards pass).

\section{Hyperparameters}

\subsection{Constants Across All Training Runs}

All training runs discussed in this work use the AdamW optimizer with $\beta_2$=0.999, and a learning rate scheduler composed of a linear warmup, followed by cosine decay. Unlike many prior works, we do not use validation-based early stopping. All graph creation uses a cutoff distance of 5.0 \AA. Tables~\ref{tab:pretrain_hparams} and \ref{tab:scd_pretrain_hparams} list the SCD pre-training hyperparameters; Tables~\ref{tab:qm9_finetune_hparams}, \ref{tab:matbench_mpgap_hparams}, \ref{tab:lba_hparams} and \ref{tab:md17_hparams} list the fine-tuning hyperparameters for QM9, Matbench, LBA and MD17.

\begin{table}[t]
\centering
\caption{Base SCD pre-training hyperparameters: applied in all SCD training runs }
\label{tab:pretrain_hparams}
\begin{tabular}{lc}
\toprule
\textbf{Hyperparameter} & \textbf{Value} \\
\midrule
Warmup steps        & 10{,}000 \\
Learning rate       & 0.005 \\
$\beta_1$           & 0.9 \\
Weight decay        & 0.05 \\
DropPath            & 0.1 \\
Corrupt noise $\sigma$ & 0.04 \\
Regularization noise $\sigma$ & 0.005 \\
\bottomrule
\end{tabular}
\end{table}

\begin{table}[t]
\centering
\caption{SCD Pre-training: dataset and model specific hyperparameters.}
\label{tab:scd_pretrain_hparams}
\begin{tabular}{lccccccc}
\toprule
Backbone            & CT  & CGT & CT & CT & CT & CT & CT \\
Dataset             & PCQ & PCQ & GEOM10 & AMP20 & SAIR & ALL & OMol25 \\
Total steps         & 800k & 800k & 800k & 800k & 800k & 1.2M & 800k \\
Batch size / GPU    & 128 & 32 & 48 & 142 & 12 & 54 & 48 \\
\# GPUs (A100)      & 4 & 16 & 16 & 4 & 8 & 16 & 16 \\
Effective batch size& 512 & 512 & 768 & 568 & 96 & 864 & 768 \\
\midrule
\multicolumn{8}{l}{\textbf{Materials augmentations}} \\
Cell repeat iters   & -- & -- & -- & 2 & -- & 2 & -- \\
$p_{\text{cell repeat}}$ & -- & -- & -- & 0.5 & -- & 0.5 & -- \\
\bottomrule
\end{tabular}
\end{table}

\begin{table}[t]
\centering
\caption{QM9 fine-tuning hyperparameters for different target groups.}
\label{tab:qm9_finetune_hparams}
\begin{tabular}{lccc}
\toprule
\textbf{Hyperparameter} &
\textbf{HOMO, LUMO, Gap} &
\textbf{$U_0$, $U$, $H$, $G$} &
\textbf{$\alpha$, ZPVE, $C_v$} \\
\midrule
Batch size                 & 128   & 32    & 128   \\
Learning rate              & 5e{-4} & 3e{-4} & 3e{-4} \\
Total steps                & 300k  & 500k  & 400k  \\
Warmup steps               & 10k   & 10k   & 10k   \\
$\beta_1$                  & 0.995 & 0.995 & 0.95  \\
Weight decay               & 0.01  & 0.01  & 0.01  \\
DropPath (init / final)    & 0.15 / 0.1 & 0.15 / 0.0 & 0.15 / 0.01 \\
Reg. noise $\sigma$        & 0.005 & 0.0 & 0.005 \\
Denoise loss weight        & 0.1   & 0.0   & 0.2   \\
Subtract ref. energy       & No    & Yes   & No    \\
Loss EMA                   & Off   & 0.05  & Off   \\
Head aggregation           & Sum   & Sum   & Sum   \\
\bottomrule
\end{tabular}
\end{table}

\begin{table}[t]
\centering
\caption{Matbench fine-tuning hyperparameters for the \texttt{mp\_gap} task.}
\label{tab:matbench_mpgap_hparams}
\begin{tabular}{lc}
\toprule
\textbf{Hyperparameter} & \textbf{Value} \\
\midrule
Batch size                 & 8 \\
Learning rate (from scratch) & 2e{-4} \\
Learning rate (fine-tuning)   & 8e{-5} \\
Total steps                & 300k \\
Warmup steps               & 10k \\
$\beta_1$                  & 0.995 \\
Weight decay               & 0.01 \\
DropPath (init / final)    & 0.15 / 0.00 \\
Reg. noise $\sigma$        & 5e{-4} \\
Denoise loss weight        & 0.01 \\
Head aggregation           & Sum \\
\bottomrule
\end{tabular}
\end{table}

\begin{table}[t]
\centering
\caption{Fine-tuning hyperparameters for the Ligand Binding Affinity (LBA) benchmark}
\label{tab:lba_hparams}
\begin{tabular}{lc}
\toprule
\textbf{Hyperparameter} & \textbf{Value} \\
\midrule
Batch size                  & 8 \\
Learning rate (from scratch) & 2e{-4} \\
Learning rate (fine-tuning)   & 1e{-4} \\
Total steps                 & 30k \\
Warmup steps                & 500 \\
$\beta_1$                   & 0.995 \\
Weight decay                & 0.05 \\
DropPath                    & 0.15 \\
Reg. noise $\sigma$         & 0.005 \\
Denoise loss weight         & 0.1 \\
Head aggregation            & Mean \\
\bottomrule
\end{tabular}
\end{table}

\begin{table}[t]
\centering
\caption{Training hyperparameters for the MD17 benchmark.}
\label{tab:md17_hparams}
\begin{tabular}{lc}
\toprule
\textbf{Hyperparameter} & \textbf{Value} \\
\midrule
Backbone                  & CT-small \\
Training set size         & 950 \\
Batch size                & 8 \\
Energy loss EMA           & 0.05 \\
Energy loss weight        & 0.8 \\
Force loss weight         & 0.2 \\
Learning rate             & 5e{-4} \\
Warmup steps              & 5k \\
Total steps               & 150k \\
$\beta_1$                 & 0.995 \\
Weight decay              & 0.01 \\
DropPath (init / final)   & 0.15 / 0.00 \\
Reg. noise $\sigma$       & 0.005 \\
Denoise loss weight       & 0.1 \\
Head aggregation          & Sum \\
\bottomrule
\end{tabular}
\end{table}

\end{document}